\documentclass[conference]{IEEEtran}
\usepackage{cite}
\usepackage{flushend}
\usepackage{stfloats}
\usepackage{multicol}
\usepackage{graphicx}
\usepackage{epstopdf}
\usepackage{setspace}
\usepackage{lettrine}
\usepackage{amsmath}
\usepackage{amsbsy}
\usepackage{mathrsfs}
\usepackage{amsfonts,amssymb}
\usepackage{algorithm}
\usepackage{algorithmic}
\usepackage{graphicx}
\usepackage{subfigure}
\usepackage{booktabs}
\usepackage{amsmath} 
\usepackage{amssymb}  

\begin{document}

%
\title{ \LARGE{Kernel Risk-Sensitive Loss: Definition, Properties and Application to Robust Adaptive Filtering}}

\author{Badong Chen, \emph{Senior Member, IEEE}, Lei Xing, Bin Xu, \emph{Member, IEEE}, Haiquan Zhao, \emph{Member, IEEE}\\
 Nanning Zheng, \emph{Fellow, IEEE}, Jos\'e C. Pr\'incipe, \emph{Fellow, IEEE}}

\maketitle

\begin{abstract}
Nonlinear similarity measures defined in kernel space, such as correntropy, can extract higher-order statistics of data and offer potentially significant performance improvement over their linear counterparts especially in non-Gaussian signal processing and machine learning. In this work, we propose a new similarity measure in kernel space, called the \emph{kernel risk-sensitive loss} (KRSL), and provide some important properties. We apply the KRSL to adaptive filtering and investigate the robustness, and then develop the MKRSL algorithm and analyze the mean square convergence performance. Compared with correntropy, the KRSL can offer a more efficient performance surface, thereby enabling a gradient based method to achieve faster convergence speed and higher accuracy while still maintaining the robustness to outliers. Theoretical analysis results and superior performance of the new algorithm are confirmed by simulation.

\end{abstract}


\let\thefootnote\relax\footnotetext{This work was supported by 973 Program (No. 2015CB351703) and National NSF of China (No. 61372152).
\par Badong Chen, Lei Xing, and Nanning Zheng are with the Institute of Artificial Intelligence and Robotics, Xi'an Jiaotong University, Xi'an, 710049, China.({chenbd; nnzheng }@mail.xjtu.edu.cn; xl2010@stu.xjtu.edu.cn).
\par Bin Xu is with the School of Automation, Northwestern Polytechnical University (NPU), Xi'an, China. (binxu@nwpu.edu.cn)
\par Haiquan Zhao is with the School of Electrical Engineering, Southwest Jiaotong University, Chengdu, China. (hqzhao@home.swjtu.edu.cn)
\par Jos\'e C. Pr\'incipe is with the Department of Electrical and Computer Engineering, University of Florida, Gainesville, FL32611 USA (principe@cnel.ufl.edu), and is also with the Institute of Artificial Intelligence and Robotics, Xi'an Jiaotong University, Xi'an, 710049, China.
}


%
\IEEEpeerreviewmaketitle
\section{Introduction}
\lettrine[lines=2]{S}{TATISTICAL} similarity measures play significant roles in signal processing and machine learning communities. In particular, the cost function for a learning machine is usually a certain similarity measure between the learned model and the data generating system. Due to their simplicity, mathematical tractability and optimality under Gaussian and linear assumptions, second-order similarity measures defined in terms of inner products, such as correlation and mean square error (MSE), have been widely used to quantify how similar two random variables (or random processes) are. However, second-order statistics cannot quantify similarity fully if the random variables are non-Gaussian distributed{\cite{principe2010information,chen2013system}}. To address the problem of modeling with non-Gaussian data (which are very common in many real-world applications), similarity measures must go beyond second-order statistics. Higher-order statistics, such as kurtosis, skewness, higher-order moments or cumulants, can be applicable for dealing with non-Gaussian data. Besides, as an alternative to the MSE, the risk-sensitive loss{\cite{jacobson1973optimal,james1994risk,liu2003output,boel2002robustness,lo2002existence}}, which quantifies the similarity by emphasizing the larger errors in an exponential form (\emph{"risk-sensitive"} means \emph{"average-of-exponential"}), has been proven to be robust to the case where the actual probabilistic model deviates from the assumed Gaussian model{\cite{boel2002robustness}}. The problem of existence and uniqueness of the risk-sensitive estimation has been studied in {\cite{lo2002existence}}. Nevertheless, the risk-sensitive loss is not robust to impulsive noises (or outliers) when utilizing a gradient-based learning algorithm, because its performance surface (as a function over parameter space) can be super-convex and the gradient may grow exponentially fast with error increasing across iterations.

\par Recent advances in \emph{information theoretic learning} (ITL) suggest that similarity measures can also be defined in a reproducing kernel Hilbert space (RKHS){\cite{principe2010information}}.The ITL costs (i.e. entropy and divergence) for adaptive system training can be directly estimated from data via a Parzen kernel estimator, which can capture higher-order statistics of data and achieve better solutions than MSE particularly in non-Gaussian and nonlinear signal processing {\cite{erdogmus2006linear,gokcay2002information,erdogmus2002error,erdogmus2002generalized,santamaria2002entropy,chen2010mean,chen2012survival}}. As a local similarity measure in ITL, the correntropy is defined as a generalized correlation in kernel space, directly related to the probability of how similar two random variables are in a neighborhood of the joint space {\cite{santamaria2006generalized,liu2007correntropy}}. The kernel function in correntropy is usually a Gaussian kernel, but it can be extended to generalized Gaussian functions {\cite{chen2016generalized}}. Since correntropy measures the similarity in an “observation window” (controlled by the kernel bandwidth), it provides an effective way to eliminate the adverse effects of outliers {\cite{principe2010information}}. So far correntropy has been successfully applied to develop various robust learning or adaptive algorithms {\cite{he2011maximum,he2014robust,he2011robust,singh2009using,zhao2011kernel,singh2010loss,chen2014steady,shi2014convex,chen2015convergence,chen2016efficient,zhu2016correntropy}}. However, the performance surface of the correntropic loss (C-Loss) is highly non-convex, which can be very sharp around the optimal solution while extremely flat at a region far away from the optimal solution, and this may result in poor convergence performance in adaptation {\cite{syed2014optimization}}.

\par In this paper, we define a new similarity measure in kernel space, called the \emph{kernel risk-sensitive loss} (KRSL), which inherits the original form of risk-sensitive loss but is defined in RKHS by means of kernel trick. The performance surface of KRSL is bounded but can be more "convex" than that of C-Loss, leading to a faster convergence speed and higher  solution accuracy while maintaining the robustness to outliers. Besides the kernel bandwidth, an extra free parameter, namely the \emph{risk-sensitive parameter}, is introduced to control the shape of the performance surface. Further, we apply the KRSL to develop a new robust adaptive filtering algorithm, referred to in this work as the \emph{minimum kernel risk-sensitive loss} (MKRSL) algorithm, which can outperform existing methods, including the recently proposed \emph{generalized maximum correntropy criterion} (GMCC) algorithm {\cite{chen2016generalized}}. A brief version of this work was presented at 2015 IEEE International Conference on Digital Signal Processing {\cite{chen2015risk}}.

\par The rest of the paper is organized as follows. In section II, after briefly reviewing some background on similarity measures in kernel space, we define the KRSL, and present some important properties. In section III, we apply the proposed KRSL to adaptive filtering and analyze the robustness, and develop the MKRSL algorithm and present the mean square convergence performance. In section IV, we carry out Monte Carlo simulations to confirm the theoretical results and demonstrate the superior performance of the new algorithm. Finally, we give the conclusion in section V.

\section{KERNEL RISK-SENSITIVE LOSS}
\subsection{Background on Similarity Measures in Kernel Space}
The kernel methods have been widely applied in domains of machine learning and signal processing {\cite{vapnik1998statistical,scholkopf2002learning,suykens2002weighted}}.Let $\kappa (x,y)$ be a continuous, symmetric and positive definite Mercer kernel defined over $\pmb{\mathscr{X}} \times \pmb{\mathscr{X}}$.Then the nonlinear mapping $\Phi (x) = \kappa (x,.)$ transforms the data $x$ from the input space to a functional Hilbert space, namely a reproducing kernel Hilbert space (RKHS)$\textbf{H}$, satisfying $f(x) = {\left\langle {\kappa (x,.),f} \right\rangle _\textbf{H}}$ , $\forall f \in \textbf{H}$ , where ${\left\langle {.,.} \right\rangle _\textbf{H}}$ denotes the inner product in $\textbf{H}$. In particular, we have
\begin{equation}
\kappa (x,y) = {\left\langle {\Phi (x),\Phi (y)} \right\rangle _\textbf{H}} = {\left\langle {\kappa (x,.),\kappa (y,.)} \right\rangle _\textbf{H}}
\end{equation}
There is a close relationship between kernel methods and information theoretic learning (ITL) {\cite{principe2010information}}. Most ITL cost functions, when estimated using the Parzen kernel estimator, can be expressed in terms of inner products in kernel space $\textbf{H}$. For example, the Parzen estimator of the \emph{quadratic information potential} (QIP) from samples $x(1),x(2), \cdots ,x(N) \in {\rm{\mathbb{R}}}$ can be obtained as {\cite{principe2010information}}
\begin{equation}
QIP(X) = \frac{1}{{2{N^2}\sqrt \pi  \sigma }}\sum\limits_{i = 1}^N {\sum\limits_{j = 1}^N {{\kappa _{\sqrt 2 \sigma }}\left( {x(i) - x(j)} \right)} }
\end{equation}
where ${\kappa _\sigma }(.)$ denotes the translation-invariant Gaussian kernel with bandwidth $\sigma$ , given by
\begin{equation}
{\kappa _\sigma }(x - y) = \exp \left( { - \frac{{{{(x - y)}^2}}}{{2{\sigma ^2}}}} \right)
\end{equation}
Then we have
\begin{small}
\begin{equation}
\begin{aligned}
QIP(X) &= \frac{1}{{2\sqrt \pi  \sigma }}\left\| {\frac{1}{N}\sum\limits_{i = 1}^N {\Phi (x(i))} } \right\|_\textbf{H}^2 \\
&= \frac{1}{{2\sqrt \pi  \sigma }}{\left\langle {\frac{1}{N}\sum\limits_{i = 1}^N {\Phi (x(i))} ,\frac{1}{N}\sum\limits_{j = 1}^N {\Phi (x(j))} } \right\rangle _\textbf{H}}
\end{aligned}
\end{equation}
\end{small}
where ${\left\| . \right\|_\textbf{H}}$ stands for the norm in \textbf{H}. Thus, the estimated QIP represents the squared mean of the transformed data in kernel space.  In this work, we only consider the Gaussian kernel of (3), but most of the results can be readily extended to other Mercer kernels.

\par The intrinsic link  between ITL and kernel methods  enlightens researchers to define new similarity measures in kernel space. As a local similarity measure in ITL, the correntropy between two random variables $X$ and $Y$, is defined by {\cite{liu2007correntropy,chen2012maximum}}
\begin{equation}
V(X,Y) = \textbf{E}\left[ {{\kappa _\sigma }(X - Y)} \right] = \int {{\kappa _\sigma }(x - y)d{F_{XY}}(x,y)}
\end{equation}
where $\textbf{E}[.]$ denotes the expectation operator, and ${F_{XY}}(x,y)$ is the joint distribution function of $(X,Y)$. Of course the correntropy $V(X,Y)$ can be expressed in terms of inner product as
\begin{equation}
V(X,Y) = \textbf{E}\left[ {{{\left\langle {\Phi (X),\Phi (Y)} \right\rangle }_\textbf{H}}} \right]
\end{equation}
which is a correlation measure in kernel space. It has been shown that correntropy is directly related to the probability of how similar two random variables are in an ``observation window"  controlled by the kernel bandwidth $\sigma$ {\cite{liu2007correntropy}}. In a similar way, one can define other similarity measures in terms of inner products in kernel space, such as the centered correntropy, correntropy coefficient and correntropic loss (C-Loss) {\cite{principe2010information}}. Three similarity measures in kernel space and their linear counterparts in input space are presented in Table 1. Similarity measures in kernel space are able to extract higher-order statistics of data and offer potentially significant performance improvement over their linear counterparts especially in non-Gaussian signal processing and machine learning {\cite{principe2010information}}.
\begin{table*}[]
\renewcommand\arraystretch{1.5}
\setlength{\abovecaptionskip}{0pt}
\setlength{\belowcaptionskip}{5pt}
\centering
\caption{Similarity measures in kernel space and their linear counterparts in input space}
\begin{tabular}{c|c|c|c}
\toprule
\multicolumn{2}{c|}{Similarity measures in kernel space}&\multicolumn{2}{|c}{Linear counterparts in input space}\\
\hline
Centered Correntropy&$\begin{array}{l}
U(X,Y) = \textbf{E}\left[ {{\kappa _\sigma }(X - Y)} \right] \\
- {\textbf{E}_X}{\textbf{E}_Y}\left[ {{\kappa _\sigma }(X - Y)} \right]
\end{array}$&Covariance&$Cov\left( {X,Y} \right) = \textbf{E}\left[ {XY} \right] - \textbf{E}\left[ X \right]\textbf{E}\left[ Y \right]$\\
\hline
Correntropy Coefficient&$\eta \left( {X,Y} \right) = \frac{{U(X,Y)}}{{\sqrt {U(X,X)U(Y,Y)} }}$ & Correlation Coefficient & $Corr\left( {X,Y} \right) = \frac{{Cov\left( {X,Y} \right)}}{{\sqrt {Cov\left( {X,X} \right)Cov\left( {Y,Y} \right)} }}$\\
\hline
Correntropic Loss&${C_{loss}}\left( {X,Y} \right) = 1 - \textbf{E}\left[ {{\kappa _\sigma }(X - Y)} \right]$&Mean Square Error&$MSE\left( {X,Y} \right) = \textbf{E}\left[ {{{\left( {X - Y} \right)}^2}} \right]$\\
\bottomrule
\end{tabular}
\end{table*}

\subsection{Kernel Risk-Sensitive Loss}
Correntropy is a local similarity measure that is little influenced by large outliers. This desirable feature makes it possible for researchers to develop robust learning algorithms using correntropy as the cost function. For example, the supervised learning problem can be solved by maximizing the correntropy (or equivalently, minimizing the C-Loss) between the model output and the desired response. This learning principle is referred to in the literature as the \emph{maximum correntropy criterion}(MCC) {\cite{principe2010information,chen2013system}}.However, the C-Loss performance surface can be highly non-convex, with steep slopes around the optimal solution while extremely flat areas away from the solution, leading to slow convergence as well as poor accuracy. This situation can be improved by choosing a larger kernel bandwidth, but with the kernel bandwidth increasing the robustness will decrease significantly when outliers occur. To achieve a better performance surface, we define in this work a new similarity measure in kernel space, called the \emph{kernel risk-sensitive loss} (KRSL). The superiority of the performance surface of KRSL will be demonstrated in the next section.

\par Given two random variables $X$ and $Y$, the KRSL is defined by
\begin{small}
\begin{equation}
\begin{aligned}
{L_\lambda }(X,Y) &= \frac{1}{\lambda }\textbf{E}\left[ {\exp \left( {\lambda \left( {1 - {\kappa _\sigma }(X - Y)} \right)} \right)} \right]\\
{\rm{              }} \\
&= \frac{1}{\lambda }\int {\exp \left( {\lambda \left( {1 - {\kappa _\sigma }(x - y)} \right)} \right)d{F_{XY}}(x,y)}
\end{aligned}
\end{equation}
\end{small}with $\lambda  > 0$ being the \emph{risk-sensitive parameter}. The above KRSL can also be expressed as
\begin{equation}
{L_\lambda }(X,Y) = \frac{1}{\lambda }\textbf{E}\left[ {\exp \left( {\lambda \left( {\frac{1}{2}\left\| {\Phi (X) - \Phi (Y)} \right\|_H^2} \right)} \right)} \right]
\end{equation}
which takes the same form as that of the traditional risk-sensitive loss {\cite{boel2002robustness,lo2002existence}}, but defined in different spaces.
\par In most practical situations, the joint distribution of $X$ and $Y$ is unknown, but only a finite number of samples $\left\{ {x(i),y(i)} \right\}_{i = 1}^N$ are available. In these cases, however, one can compute an approximation, called \emph{empirical KRSL}, by approximating the expectation by an average over \emph{$N$} samples:
\begin{equation}
{\hat L_\lambda }(X,Y) = \frac{1}{{N\lambda }}\sum\limits_{i = 1}^N {\exp \left( {\lambda \left( {1 - {\kappa _\sigma }(x(i) - y(i))} \right)} \right)}
\end{equation}
The empirical KRSL also defines a ``distance" between the vectors $\textbf{X} = {\left[ {x(1),x(2), \cdots ,x(N)} \right]^T}$ and $\textbf{Y} = {\left[ {y(1),y(2), \cdots ,y(N)} \right]^T}$ . In this work, we also denote ${\hat L_\lambda }(X,Y)$ by ${\hat L_\lambda }(\textbf{X},\textbf{Y})$ when no confusion arises.

\subsection{Properties}
In the following, we present some important properties of the proposed KRSL.
\par \emph{Property 1}: ${L_\lambda }(X,Y)$ is symmetric, that is ${L_\lambda }(X,Y) = {L_\lambda }(Y,X)$.
\par \emph{Proof}: Straightforward since ${\kappa _\sigma }(X - Y) = {\kappa _\sigma }(Y - X)$.
\par \emph{Property 2}: ${L_\lambda }(X,Y)$ is positive and bounded: $\frac{1}{\lambda } \le {L_\lambda }(X,Y) \le \frac{1}{\lambda }\exp \left( \lambda  \right)$, and it reaches its minimum if and only if $X = Y$.
\par \emph{Proof}: Straightforward since $0 < {\kappa _\sigma }(X - Y) \le 1$, and ${\kappa _\sigma }(X - Y) = 1$ if and only if $X=Y$.
\par \emph{Property 3}: As $\lambda$ is small enough, it holds that ${L_\lambda }(X,Y) \approx \frac{1}{\lambda } + {C_{loss}}\left( {X,Y} \right)$.
\par \emph{Proof}: For $\lambda$ small enough, we have $\exp \left( {\lambda \left( {1\!-\!{\kappa _\sigma }(X \!-\! Y)} \right)} \right)\!\approx\! 1\!+\!\lambda \left( {1 \!-\! {\kappa _\sigma }(X \!-\! Y)} \right)$, and it follows that
\begin{equation}
\begin{aligned}
{L_\lambda }(X,Y) &= \frac{1}{\lambda }\textbf{E}\left[ {\exp \left( {\lambda \left( {1 - {\kappa _\sigma }(X - Y)} \right)} \right)} \right]\\
&{\rm{              }} \approx \frac{1}{\lambda }\textbf{E}\left[ {1 + \lambda \left( {1 - {\kappa _\sigma }(X - Y)} \right)} \right]\\
&{\rm{              }} = \frac{1}{\lambda } + \textbf{E}\left[ {1 - {\kappa _\sigma }(X - Y)} \right]\\
&{\rm{              }} = \frac{1}{\lambda } + {C_{loss}}\left( {X,Y} \right)
\end{aligned}
\end{equation}
\par \emph{Property 4}: As $\sigma$ is large enough, it holds that ${L_\lambda }(X,Y) \approx \frac{1}{\lambda } + \frac{1}{{2{\sigma ^2}}}\textbf{E}\left[ {{{\left( {X - Y} \right)}^2}} \right]$.
\par \emph{Proof}: Due to $\exp (x) \approx 1 + x$ for small enough, as $\sigma$ is large enough, we have
\begin{equation}
\begin{aligned}
{L_\lambda }(X,Y) &= \frac{1}{\lambda }\textbf{E}\left[ {\exp \left( {\lambda \left( {1 - \exp \left( { - \frac{{{{\left( {X - Y} \right)}^2}}}{{2{\sigma ^2}}}} \right)} \right)} \right)} \right]\\
&{\rm{              }} \approx \frac{1}{\lambda }\textbf{E}\left[ {\exp \left( {\lambda \left( {\frac{{{{\left( {X - Y} \right)}^2}}}{{2{\sigma ^2}}}} \right)} \right)} \right]\\
&{\rm{              }} \approx \frac{1}{\lambda }\textbf{E}\left[ {1 + \lambda \left( {\frac{{{{\left( {X - Y} \right)}^2}}}{{2{\sigma ^2}}}} \right)} \right]\\
&{\rm{              }} = \frac{1}{\lambda } + \frac{1}{{2{\sigma ^2}}}\textbf{E}\left[ {{{\left( {X - Y} \right)}^2}} \right]
\end{aligned}
\end{equation}
\par \emph{Remark 1}: According to Property 3 and 4, the KRSL will be, approximately, equivalent to the C-Loss as $\lambda$ is small enough, and equivalent to the MSE when $\sigma$ is large enough. Thus the C-Loss and MSE can be viewed as two extreme cases of the KRSL.
\par \emph{Property 5}: Let $\emph{\textbf{e}} = \textbf{X} - \textbf{Y} = {\left[ {e(1),e(2), \cdots ,e(N)} \right]^T}$, where $e(i) = x(i) - y(i)$. Then the empirical KRSL ${\hat L_\lambda }(\textbf{X},\textbf{Y})$ as a function of $\emph{\textbf{e}}$ is convex at any point satisfying ${\left\| \emph{\textbf{e}} \right\|_\infty } = \mathop {\max }\limits_{i = 1,2, \cdots N} \left| {e(i)} \right| \le \sigma $ .
\par \emph{Proof}: Since ${\hat L_\lambda }(\textbf{X},\textbf{Y}) = \frac{1}{{N\lambda }}\sum\limits_{i = 1}^N {\exp \left( {\lambda \left( {1 - {\kappa _\sigma }(e(i))} \right)} \right)} $ , the Hessian matrix of ${\hat L_\lambda }(\textbf{X},\textbf{Y})$ with respect to $\emph{\textbf{e}}$ can be derived as
\begin{equation}
{H_{{{\hat L}_\lambda }(\textbf{X},\textbf{Y})}}\left( \emph{\textbf{e}} \right) = \left[ {\frac{{{\partial ^2}{{\hat L}_\lambda }(\textbf{X},\textbf{Y})}}{{\partial e(i)\partial e(j)}}} \right] = diag\left[ {{\gamma _1},{\gamma _2}, \cdots ,{\gamma _N}} \right]
\end{equation}
where
\begin{equation}
{\gamma _i} = {\xi _i}\left( {\frac{\lambda }{{{\sigma ^2}}}\exp \left( { - \frac{{{e^2}(i)}}{{2{\sigma ^2}}}} \right){e^2}(i) + 1 - \frac{1}{{{\sigma ^2}}}{e^2}(i)} \right)
\end{equation}
with ${\xi _i} = \frac{1}{{N{\sigma ^2}}}\exp \left( {\lambda \left( {1 - \exp \left( { - \frac{{{e^2}(i)}}{{2{\sigma ^2}}}} \right)} \right)} \right)\exp \left( { - \frac{{{e^2}(i)}}{{2{\sigma ^2}}}} \right) > 0$. Thus we have ${H_{{{\hat L}_\lambda }(\textbf{X},\textbf{Y})}}\left( \emph{\textbf{e}} \right) > 0$ if $\mathop {\max }\limits_{i = 1,2, \cdots N} \left| {e(i)} \right| \le \sigma $ . This completes the proof.

\par \emph{Property 6}: Given any point $\emph{\textbf{e}}$ with ${\left\| \emph{\textbf{e}} \right\|_\infty } > \sigma $, the empirical KRSL ${\hat L_\lambda }(\textbf{X},\textbf{Y})$ will be convex at $\emph{\textbf{e}}$ if the risk-sensitive parameter $\lambda$ is larger than a certain value..
\par \emph{Proof}: From (13), we have ${\gamma _i} \ge 0$ if one of the following conditions is satisfied: i)$\left| {e(i)} \right| \le \sigma $; ii)$\left| {e(i)} \right| > \sigma $ and $\lambda  \ge \left( {\frac{{{e^2}(i) - {\sigma ^2}}}{{{e^2}(i)}}} \right)\exp \left( {\frac{{{e^2}(i)}}{{2{\sigma ^2}}}} \right)$.
Therefore, we have ${H_{{{\hat L}_\lambda }(\textbf{X},\textbf{Y})}}\left( \emph{\textbf{e}} \right) \ge 0$ if
\begin{equation}
\lambda  \ge \mathop {\max }\limits_{i= 1, \cdots ,N\atop| {e(i)} | > \sigma }\left \{\left [\frac{e^{2}(i)-\sigma ^{2}}{e^{2}(i)}\right ]\mathop {\exp }\left [ \frac{e^{2}(i)}{2\sigma ^{2}}\right ]\right \}
\end{equation}
This complete the proof.

\par \emph{Remark 2}: According to Property 5 and 6, the empirical KRSL as a function of \emph{\textbf{e}} is convex at any point satisfying ${\left\| \emph{\textbf{e}} \right\|_\infty } \le \sigma $. For the case ${\left\| \emph{\textbf{e}} \right\|_\infty } > \sigma $ , the empirical KRSL can still be convex at \emph{\textbf{e}} if the risk-sensitive parameter $\lambda$ is larger than a certain value. In fact, the parameter $\lambda$ controls the convex range, and a larger $\lambda$ results in a larger convex range in general.

\par \emph{Property 7}: As $\sigma  \to \infty $ (or $x(i) \to 0,i = 1, \cdots ,N$ ), it holds that
\begin{equation}
{\hat L_\lambda }(\textbf{X},\textbf{0}) \approx \frac{1}{{2{\sigma ^2}}}\left\| \textbf{X} \right\|_2^2 + \frac{1}{\lambda }
\end{equation}
where \textbf{0} denotes an $N$-dimensional zero vector.

\par \emph{Proof}: Since $\exp (x) \approx 1 + x$ as $x \to 0$, as $\sigma$ is large enough, we have
\begin{small}
\begin{equation}
\begin{aligned}
{{\hat L}_\lambda }(\textbf{X},\textbf{0}) &= \frac{1}{{N\lambda }}\sum\limits_{i = 1}^N {\exp \left( {\lambda \left( {1 - {\kappa _\sigma }(x(i))} \right)} \right)} \\
&{\rm{              }} \approx \frac{1}{{N\lambda }}\sum\limits_{i = 1}^N {\exp \left( {\lambda \left( {1 - \left( {1 - \frac{{{x^2}(i)}}{{2{\sigma ^2}}}} \right)} \right)} \right)} \\
&{\rm{              }} = \frac{1}{{N\lambda }}\sum\limits_{i = 1}^N {\exp \left( {\lambda \frac{{{x^2}(i)}}{{2{\sigma ^2}}}} \right)} \\
&{\rm{              }} \approx \frac{1}{{N\lambda }}\sum\limits_{i = 1}^N {\left[ {1 + \lambda \frac{{{x^2}(i)}}{{2{\sigma ^2}}}} \right]} \\
&{\rm{              }} = \frac{1}{\lambda } + \frac{1}{{2{\sigma ^2}}}\frac{1}{N}\sum\limits_{i = 1}^N {{x^2}(i)} \\
&{\rm{             }} = \frac{1}{{2{\sigma ^2}}}\left\| \textbf{X} \right\|_2^2 + \frac{1}{\lambda }
\end{aligned}
\end{equation}
\end{small}
\par \emph{Property 8}: Assume that $\left| {{x_i}} \right| > \delta $, $\forall i:{x_i} \ne 0$, where $\delta$ is a small positive number. As  $\sigma  \to 0 + $, minimizing the empirical KRSL ${\hat L_\lambda }(\textbf{X},\textbf{0})$ will be, approximately, equivalent to minimizing the ${l_0}$-norm of $\textbf{X}$, that is
\begin{equation}
\mathop {\min }\limits_{\textbf{X} \in \Omega } {\hat L_\lambda }(\textbf{X},\textbf{0}) \sim \mathop {\min }\limits_{\textbf{X} \in \Omega } {\left\| \textbf{X} \right\|_0},\;\;as\;\; \sigma  \to 0 +
\end{equation}
where $\Omega $ denotes a feasible set of $\textbf{X}$.

\par \emph{Proof}: Let ${\textbf{X}_0}$ be the solution obtained by minimizing ${\left\| \textbf{X} \right\|_0}$ over $\Omega $ and ${ \textbf{X} _\textbf{L}}$ the solution achieved by minimizing ${\hat L_\lambda }(\textbf{X},\textbf{0})$. Then ${\hat L_\lambda }({\textbf{X}_L},\textbf{0}) \le {\hat L_\lambda }({\textbf{X}_0},\textbf{0})$, and
\begin{equation}
\begin{aligned}
&\sum\limits_{i = 1}^N {\left[ {\exp \left( {\lambda \left( {1 - {\kappa _\sigma }\left( {{{({\textbf{X}_L})}_i}} \right)} \right)} \right) - \exp (\lambda )} \right]}\\
&\le \sum\limits_{i = 1}^N {\left[ {\exp \left( {\lambda \left( {1 - {\kappa _\sigma }\left( {{{({\textbf{X}_0})}_i}} \right)} \right)} \right) - \exp (\lambda )} \right]}
\end{aligned}
\end{equation}
where ${({\textbf{X}_L})_i}$ denotes the $i$th component of ${\textbf{X}_L}$. It follows that
\begin{equation}
\begin{aligned}
&\left( {1 - \exp (\lambda )} \right)\left( {N - {{\left\| {{\textbf{X}_L}} \right\|}_0}} \right) \\
&+ \sum\limits_{i = 1,{{({\textbf{X}_L})}_i} \ne 0}^N {\left[ {\exp \left( {\lambda \left( {1 - {\kappa _\sigma }\left( {{{({\textbf{X}_L})}_i}} \right)} \right)} \right) - \exp (\lambda )} \right]} \\
&{\rm{             }} \le \left( {1 - \exp (\lambda )} \right)\left( {N - {{\left\| {{\textbf{X}_0}} \right\|}_0}} \right) \\
&+ \sum\limits_{i = 1,{{({\textbf{X}_0})}_i} \ne 0}^N {\left[ {\exp \left( {\lambda \left( {1 - {\kappa _\sigma }\left( {{{({\textbf{X}_0})}_i}} \right)} \right)} \right) - \exp (\lambda )} \right]}
\end{aligned}
\end{equation}
Hence
\begin{equation}
\begin{aligned}
{\left\| {{\textbf{X}_L}} \right\|_0} \!\!\!-\!\!\! {\left\| {{\textbf{X}_0}} \right\|_0} &\le \frac{{\sum\limits_{i = 1,{{({\textbf{X}_0})}_i} \ne 0}^N {\left[ {\exp \left( {\lambda \left( {1 - {\kappa _\sigma }\left( {{{({\textbf{X}_0})}_i}} \right)} \right)} \right) - \exp (\lambda )} \right]} }}{{\exp (\lambda ) - 1}} \\
&- \frac{{\sum\limits_{i = 1,{{({\textbf{X}_L})}_i} \ne 0}^N {\left[ {\exp \left( {\lambda \left( {1 - {\kappa _\sigma }\left( {{{({\textbf{X}_L})}_i}} \right)} \right)} \right) - \exp (\lambda )} \right]} }}{{\exp (\lambda ) - 1}}
\end{aligned}
\end{equation}
Since $\left| {{x_i}} \right| > \delta $, $\forall i:{x_i} \ne 0$ , as $\sigma  \to 0 + $ the right hand side of (20) will approach zero. Thus, if $\sigma$ is small enough, it holds that
\begin{equation}
{\left\| {{\textbf{X}_0}} \right\|_0} \le {\left\| {{\textbf{X}_L}} \right\|_0} \le {\left\| {{\textbf{X}_0}} \right\|_0} + \varepsilon
\end{equation}
where $\varepsilon $ is a small positive number arbitrarily close to zero. This completes the proof.
\par \emph{Remark 3}: According to Property 7 and 8, the empirical KRSL ${\hat L_\lambda }(\textbf{X},\textbf{0})$ behaves like a squared ${L_2}$ norm of $\textbf{X}$ when kernel bandwidth $\sigma$ is very large, and like an ${L_0}$ norm of $\textbf{X}$ when $\sigma$ is very small. Similar properties also hold for the empirical C-Loss (or \emph{correntropy induced metric}, CIM) {\cite{liu2007correntropy}}.

\begin{figure*}[htbp]
\setlength{\abovecaptionskip}{0pt}
\setlength{\belowcaptionskip}{0pt}
\centering
\subfigure[]{
\includegraphics[width=3.0in,height=2.4in]{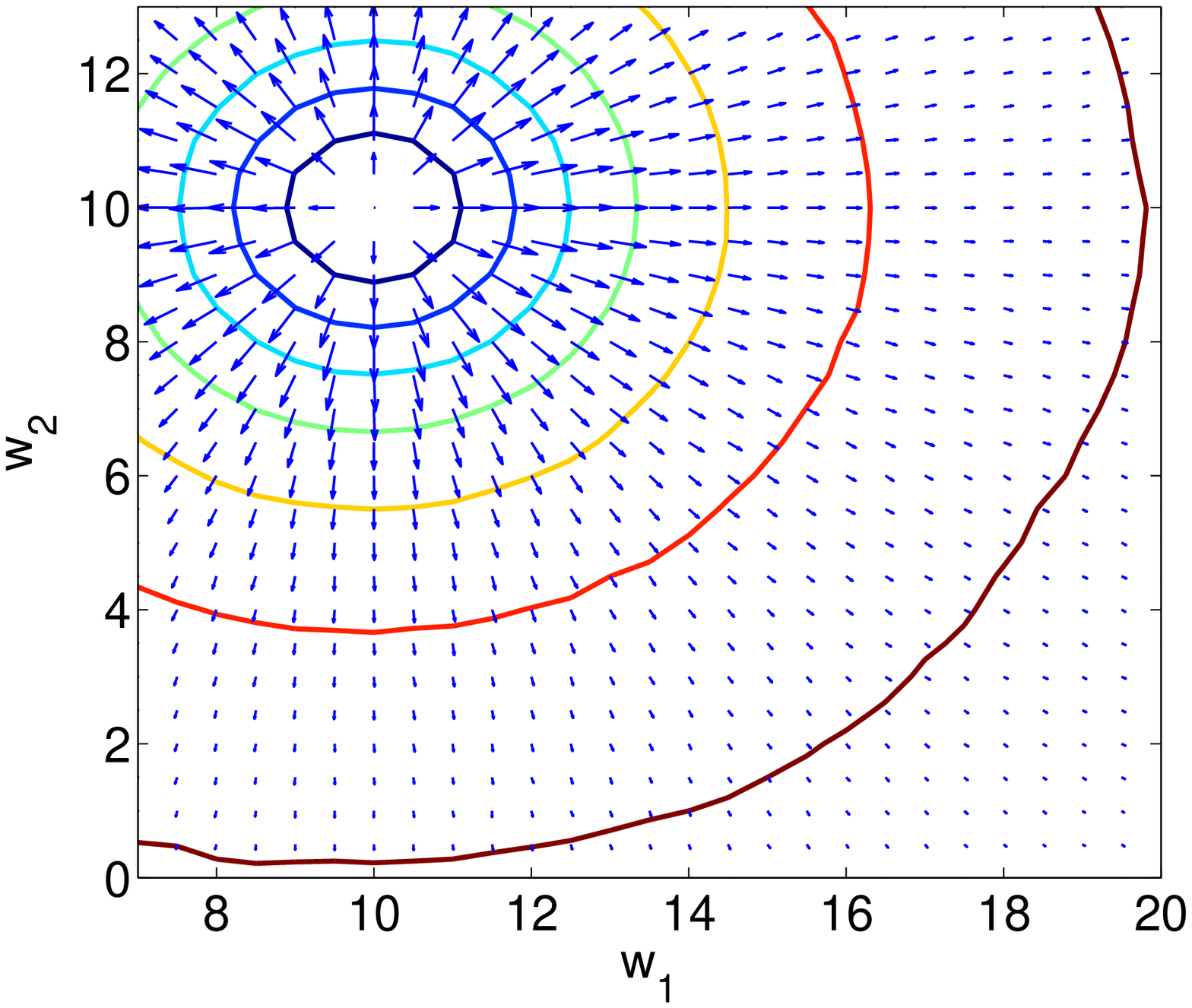}}
\subfigure[]{
\includegraphics[width=3.0in,height=2.4in]{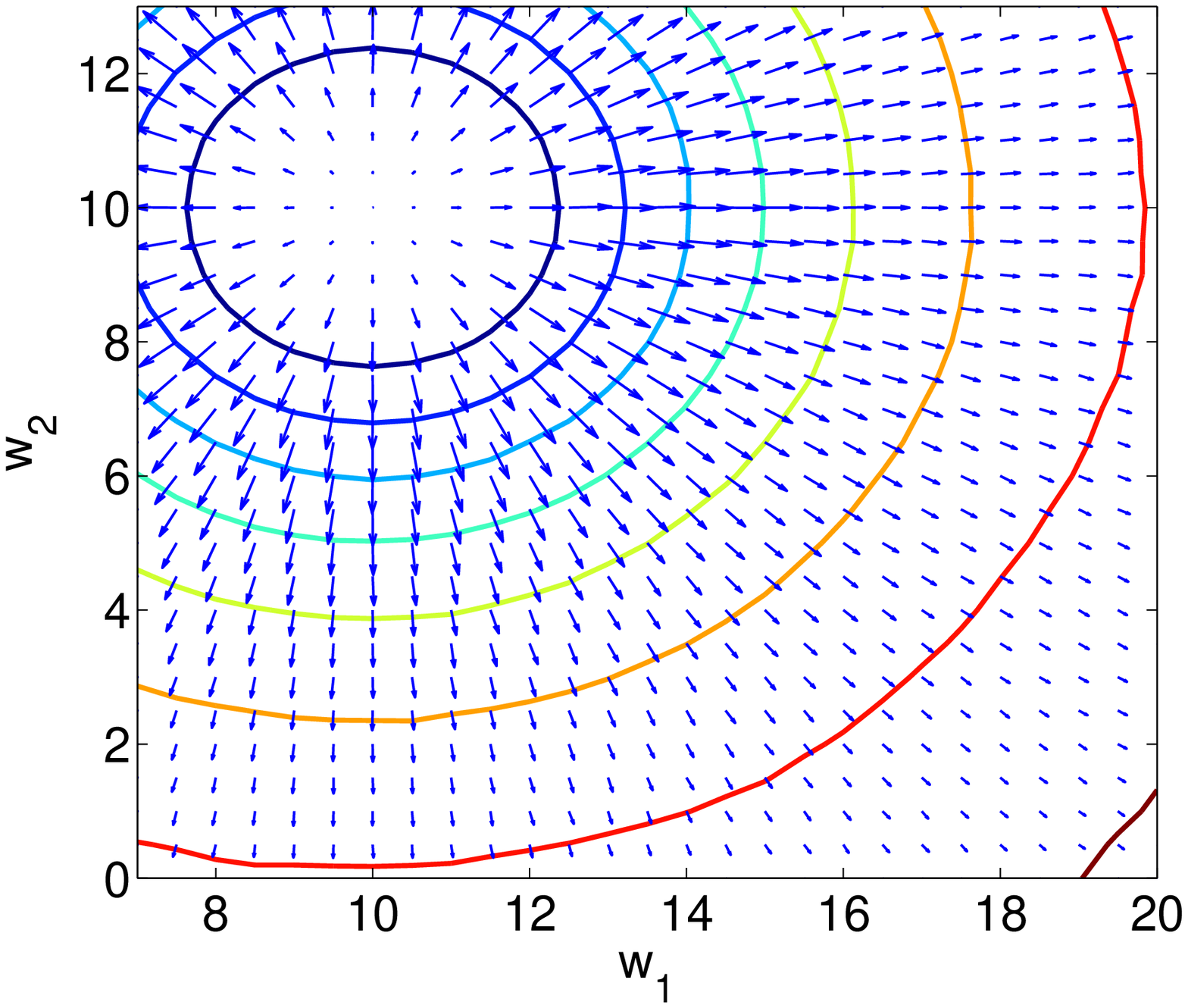}}
\caption{Contours and gradients plots of the performance surfaces: (a) C-Loss; (b) KRSL}
\label{fig1}
\end{figure*}


\section{APPLICATION TO ADAPTIVE FILTERING}
\subsection{Performance Surface}
Consider the identification of an FIR system:
\begin{equation}
d(i) = W_0^TX(i) + v(i)
\end{equation}
where $d(i) \in \mathbb{R}$ denotes an observed response at time $i$ ,${W_0} \in \mathbb{R}^{m}$ is an unknown weight vector to be estimated, $X(i) = {\left[ {x(i - m + 1), \cdots ,x(i)} \right]^T}$ is the input vector (known value), and $v(i)$ stands for an additive noise (usually independent of the input). Let ${W} \in \mathbb{R}^{m}$ be the estimated value of the weight vector. Then the KRSL cost (as a function of $W$ is also referred to as the \emph{performance surface}) is
\begin{equation}
\begin{aligned}
{J_{KRSL}}(W) &= \frac{1}{{N\lambda }}\sum\limits_{i = 1}^N {\exp \left( {\lambda \left( {1 - {\kappa _\sigma }\left( {e(i)} \right)} \right)} \right)} \\
&{\rm{             }} = \frac{1}{{N\lambda }}\sum\limits_{i = 1}^N {\exp \left( {\lambda \left( {1 - {\kappa _\sigma }\left( {d(i) - {W^T}X(i)} \right)} \right)} \right)}
\end{aligned}
\end{equation}
with $e(i) = d(i) - {W^T}X(i)$ being the error at time $i$ and $N$ the number of samples. The optimal solution can be solved by minimizing the cost function ${J_{KRSL}}(W)$. This optimization principle is called in this paper the \emph{minimum kernel risk-sensitive loss} (MKRSL) criterion. The following theorem holds.

\par \emph{Theorem  1 (Optimal Solution)}: The optimal solution under the MKRSL criterion satisfies
\begin{equation}
{W_{MKRSL}} \!\!=\!\!\! {\left[ {\sum\limits_{i = 1}^N {h\left( {e(i)} \right)} X(i)X{{(i)}^T}}\!\! \right]^{ - 1}}\!\!\left[ {\sum\limits_{i = 1}^N {h\left( {e(i)} \right)} d(i)X(i)} \right]
\end{equation}
where\\
$h(e(i)) \!\!=\!\! \exp \!\left(\! {\lambda \!\left( \!{1 \!-\! {\kappa _\sigma }\left( {d(i) \!\!-\! {W^T}X(i)} \right)} \right)} \right){\kappa _\sigma }\!\left(\! {d(i) \!\!-\! {W^T}X(i)} \!\right)$, provided that the matrix $\sum\limits_{i = 1}^N {h\left( \!{e(i)} \!\right)\!} X(i)X{(i)^T}$ is invertible.

\par \emph{Proof}: It is easy to derive
\begin{equation}
\begin{aligned}
&\frac{\partial }{{\partial W}}{J_{KRSL}}(W) = 0\\
&\Rightarrow \sum\limits_{i = 1}^N {\exp \left( {\lambda \left( {1 - {\kappa _\sigma }\left( {d(i) - {W^T}X(i)} \right)} \right)} \right)} \times\\
&{\kappa _\sigma }\left( {d(i) - {W^T}X(i)} \right)\left( {d(i) - {W^T}X(i)} \right)X(i) = 0\\
&\Rightarrow \left[ {\sum\limits_{i = 1}^N {h\left( {e(i)} \right)} X(i)X{{(i)}^T}} \right]W = \left[ {\sum\limits_{i = 1}^N {h\left( {e(i)} \right)} d(i)X(i)} \right]\\
&\Rightarrow {W_{MKRSL}} = {\left[ {\sum\limits_{i = 1}^N {h\left( {e(i)} \right)} X(i)X{{(i)}^T}} \right]^{ - 1}}\times\\
&\;\;\;\;\;\;\;\;\;\;\;\;\;\;\;\;\;\;\;\;\;\;\;\;\;\left[ {\sum\limits_{i = 1}^N {h\left( {e(i)} \right)} d(i)X(i)} \right]
\end{aligned}
\end{equation}

\emph{Remark 4}: We have $h(e(i)) \to 1$ as $\sigma  \to \infty $. In this case, the optimal solution ${W_{MKRSL}}$ will be equal to the well-known \emph{Wiener solution}. In addition, it is worth noting that the equation (24) does not provide a closed-form solution because the right hand side of (24) depends on $W$ through the error $e(i)$ .

\par Now we compare the performance surfaces of the proposed KRSL and C-Loss. For the case $m=2$ (for visualization purpose), the contours and gradients (with respect to $W$) of the performance surfaces are plotted in Fig.1, where ${W_0} = {\left[ {10,10} \right]^T}$ , $\sigma=2.0$ , $\lambda=10$ , $N=10000$ , and the input $\left\{ {x(i)} \right\}$ and noise $\left\{ {v(i)} \right\}$ are both zero-mean white Gaussian processes with unit variance. From Fig.1, one can see that the performance surface of the C-Loss is very flat (where the gradients are very small) when the estimated value is far away from the optimal solution (i.e. ${W_0}$ ), whereas it becomes very sharp near the optimal solution. For a gradient-based search algorithm, such a performance surface may lead to slow convergence speed especially when the initial estimate is far away from the optimal solution and possibly low accuracy at final stage due to misadjustments caused by large gradients near the optimal solution. By contrast, the performance surface of the KRSL has three regions: i) when the estimated value is close to the optimum, the gradients will become small to reduce the misadjustments; ii) when the estimated value is away from the optimum, the gradients will become large to speed up the convergence; iii) when the estimated value is further away from the optimum, the gradients will decrease gradually to zero to avoid big fluctuations possibly caused by large outliers. Therefore, compared with the C-Loss, the KRSL can offer potentially a more efficient solution, enabling simultaneously faster convergence and higher accuracy while maintaining the robustness to outliers.


\begin{figure*}[b!]
\begin{small}
\begin{equation}
\xi  = \frac{1}{c}\left( {\sqrt { - 2{\sigma ^2}\log \left( {1 - \frac{1}{\lambda }\log \left[ {\exp \left( {\lambda \left( {1 - \exp \left( { - \frac{{\varepsilon _v^2}}{{2{\sigma ^2}}}} \right)} \right)} \right) + \frac{{N - M}}{M}\left( {\exp (\lambda ) - 1} \right)} \right]} \right)}  + {\varepsilon _v}} \right)
\end{equation}
\end{small}
\end{figure*}

\begin{figure*}[b!]
\begin{small}
\begin{equation}
\rho  = \frac{1}{c}\left( {\beta \sqrt {\frac{{\log \left( {1 - \frac{1}{\lambda }\log \left[ {\exp \left( {\lambda \left( {1 - {{\left( {1 - \frac{1}{\lambda }\log \left[ {\exp (\lambda ) - \frac{{N - M}}{M}\left( {\exp (\lambda ) - 1} \right)} \right]} \right)}^{{1 \mathord{\left/
 {\vphantom {1 {{\beta ^2}}}} \right.
 \kern-\nulldelimiterspace} {{\beta ^2}}}}}} \right)} \right) + \frac{{N - M}}{M}\left( {\exp (\lambda ) - 1} \right)} \right]} \right)}}{{\log \left\{ {1 - \frac{1}{\lambda }\log \left[ {\exp (\lambda ) - \frac{{N - M}}{M}\left( {\exp (\lambda ) - 1} \right)} \right]} \right\}}}}  + 1} \right)
\end{equation}
\end{small}
\end{figure*}

\subsection{Robustness Analysis}
Similar to the MCC criterion, the MKRSL criterion is also robust to impulsive noises (or large outliers).
In the following, we present some theoretical results on the robustness of the MKRSL criterion. For mathematical tractability, we consider only the scalar FIR identification case ( $m=1$ ). In this case, the weight $W$ and input $X(i)$ are both scalars.
\par First, we give some notations. Let ${\varepsilon _v} > 0$ be a positive number, ${I_N} = \left\{ {1,2, \cdots ,N} \right\}$ be the sample index set, and $I\left( {{\varepsilon _v}} \right) = \left\{ {i:i \in {I_N},\left| {v(i)} \right| \le {\varepsilon _v}} \right\}$ be a subset of ${I_N}$ satisfying $\forall i \in I\left( {{\varepsilon _v}} \right)$ , $\left| {v(i)} \right| \le {\varepsilon _v}$ . In addition, the following two assumptions are made:

\par \emph{Assumption 1}: $N > \left| {I\left( {{\varepsilon _v}} \right)} \right| = M > \frac{N}{2}$, where $\left| {I\left( {{\varepsilon _v}} \right)} \right|$ denotes the cardinality of the set $I\left( {{\varepsilon _v}} \right)$ ;

\par \emph{Assumption 2}: $\exists c > 0$ such that $\forall i \in I\left( {{\varepsilon _v}} \right)$ , $\left| {X(i)} \right| \ge c$.

\par \emph{Remark 5}: The \emph{Assumption 1} means that there are $M$ ( more than $\frac{N}{2}$) samples in which the amplitudes of the additive noises satisfy $\left| {v(i)} \right| \le {\varepsilon _v}$, and $N-M$ (at least one) samples that may contain large outliers with $\left| {v(i)} \right| > {\varepsilon _v}$ (possibly $\left| {v(i)} \right| \gg {\varepsilon _v}$). The \emph{Assumption 2} is reasonable since for a finite number of samples, the minimum amplitude is non-zero in general.

\par With the above notations and assumptions, the following theorem holds:

\par \emph{Theorem 2}: if $\sigma  > \frac{{{\varepsilon _v}}}{{\sqrt { - 2\log \left\{ {1 - \frac{1}{\lambda }\log \left[ {\exp (\lambda ) - \frac{{N - M}}{M}\left( {\exp (\lambda ) - 1} \right)} \right]} \right\}} }}$, then the optimal solution ${W_{MKRSL}}$ under the MKRSL criterion satisfies $\left| {{W_{MKRSL}} - {W_0}} \right| \le \xi $, where the expression of $\xi $ is shown at the the bottom of the page.

\par \emph{Proof}: See Appendix.

\par The\! following\! corollary\! is\! a\! direct\! consequence\! of\! Theorem\! 2:

\par \emph{Corollary 1}:  If\\
$\sigma  > \frac{{{\varepsilon _v}}}{{\sqrt { - 2\log \left\{ {1 - \frac{1}{\lambda }\log \left[ {\exp (\lambda ) - \frac{{N - M}}{M}\left( {\exp (\lambda ) - 1} \right)} \right]} \right\}} }}$, then the optimal solution ${W_{MKRSL}}$ under MKRSL satisfies $\left| {{W_{MKRSL}} - {W_0}} \right| \le \rho {\varepsilon _v}$, where the expression of the constant $\rho $ is shown at the the bottom of the page, with $\beta  = \frac{\sigma }{{{\varepsilon _v}}}\sqrt { - 2\log \left\{ {1 - \frac{1}{\lambda }\log \left[ {\exp (\lambda ) - \frac{{N - M}}{M}\left( {\exp (\lambda ) - 1} \right)} \right]} \right\}} $.

\par \emph{Remark 5}: According to Corollary 1, if the kernel bandwidth $\sigma$ is larger than a certain value, the absolute value of the estimation error ${\varepsilon _{MKRSL}} = {W_{MKRSL}} - {W_0}$ will be upper bounded by $\rho {\varepsilon _v}$. If ${\varepsilon _v}$ is very small, the upper bound $\rho {\varepsilon _v}$ will also be very small, which implies that the MKRSL solution ${W_{MKRSL}}$ can be very close to the true value (${W_0}$) even in presence of $(N-M)$ outliers (whose values can be arbitrarily large), provided that there are $M$($M > {N \mathord{\left/{\vphantom {N 2}} \right.\kern-\nulldelimiterspace} 2}$) samples disturbed by small noises (bounded by ${\varepsilon _v}$).

\par For the vector case ($m > 1$), it is very difficult to derive an upper bound on the norm of the estimation error. However, we believe that the above results for scalar case explain clearly why and how the MKRSL criterion will be robust to large outliers.

\subsection{Stochastic Gradient Adaptive Algorithm}
Stochastic gradient based adaptive algorithms have been widely used in many practical applications, especially those involving online adaptation. Under the MKRSL criterion, the instantaneous cost function at time $i$ is
\begin{equation}
{\hat J_{KRSL}} = \frac{1}{\lambda }\exp \left( {\lambda \left( {1 - {\kappa _\sigma }\left( {e(i)} \right)} \right)} \right)
\end{equation}Then a stochastic gradient based adaptive filtering algorithm can be easily derived as

\begin{equation}
\begin{aligned}
&W(i + 1) = W(i) - \mu \frac{\partial }{{\partial W(i)}}{{\hat J}_{KRSL}}\\
&{\rm{            }} = W(i) + \frac{\mu }{{{\sigma ^2}}}\exp \left( {\lambda \left( {1 - {\kappa _\sigma }(e(i))} \right)} \right){\kappa _\sigma }(e(i))e(i)X(i)\\
&{\rm{            }} = W(i) + \eta \exp \left( {\lambda \left( {1 - {\kappa _\sigma }(e(i))} \right)} \right){\kappa _\sigma }(e(i))e(i)X(i)
\end{aligned}
\end{equation}where $W(i)$ denotes the estimated weight vector at time $i$, and $\eta  = \frac{\mu }{{{\sigma ^2}}}$ is the step-size parameter. We call the above algorithm the \emph{MKRSL algorithm}. In this work, we use the same abbreviation for an optimization criterion and the corresponding algorithm when no confusion can arise from the context.

\par  The MKRSL algorithm (29) can also be expressed as
\begin{equation}
W(i + 1) = W(i) + \eta (i)e(i)X(i)
\end{equation}
which is a least mean square (LMS) algorithm with a variable step-size (VSS) $\eta (i) = \eta \exp \left( {\lambda \left( {1 - {\kappa _\sigma }(e(i))} \right)} \right){\kappa _\sigma }(e(i))$ .

\par We have the following observations:

\par 1) As $\lambda  \to 0 + $, we have $\eta (i) \to \eta {\kappa _\sigma }(e(i))$. In this case, the MKRSL algorithm becomes the MCC algorithm {\cite{singh2009using,zhao2011kernel}}:
\begin{equation}
W(i + 1) = W(i) + \eta {\kappa _\sigma }(e(i))e(i)X(i)
\end{equation}

\par 2) As $\sigma  \to \infty $, we have $\eta (i) \to \eta $. In this case, the MKRSL algorithm will reduce to the original LMS algorithm (with a fixed step-size):
\begin{equation}
W(i) = W(i - 1) + \eta e(i)X(i)
\end{equation}

\par Fig. 2 shows the curves of $\eta (i)$ as a function of $e(i)$ for different values of $\lambda$ (where $\sigma  = \eta  = 2.0$). As one can see, when $\lambda=0$ (corresponding to the MCC algorithm), the step-size $\eta (i)$ will reach the maximum at the origin ($e(i) = 0$).When $\lambda>0$, however, the step-size $\eta (i)$ may reach the maximum at a location away from the origin, potentially leading to faster convergence speed and better accuracy. For any $\lambda$, the step-size $\eta (i)$ will approach zero as $\left| {e(i)} \right| \to \infty $ , which implies that the MKRSL algorithm will be insensitive  (or robust) to large errors.

\par Note that the computational complexity of the MKRSL algorithm is almost the same as the MCC algorithm. The only extra computational demand is to calculate the term $\exp \left( {\lambda \left( {1 - {\kappa _\sigma }(e(i))} \right)} \right)$.

\begin{figure}[t]
\setlength{\abovecaptionskip}{0pt}
\setlength{\belowcaptionskip}{0pt}
\centering
\includegraphics[width=3.0in,height=2.3in]{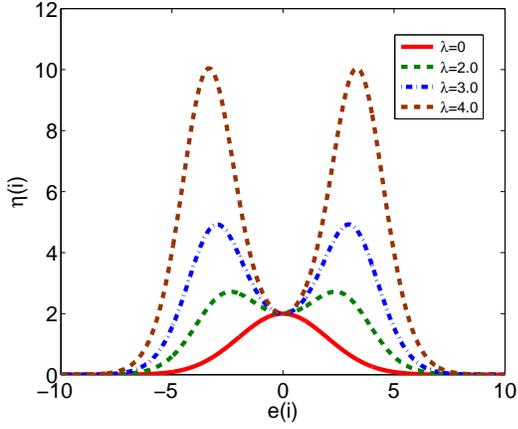}
\caption{Curves of $\eta (i)$ as a function of $e(i)$ ($\sigma  = \eta  = 2.0$ )}
\label{fig2}
\end{figure}


\subsection{Mean Square Convergence Performance}
The mean square convergence behavior is very important for an adaptive filtering algorithm. There have been extensive studies on the mean square convergence of various adaptive filtering algorithms in the literature {\cite{sayed2003fundamentals}}. The proposed MKRSL algorithm belongs to a general class of adaptive filtering algorithms {\cite{al2001adaptive,al2003transient}}:

\begin{equation}
W(i + 1) = W(i) + \eta f\left( {e(i)} \right)X(i)
\end{equation}
where $f\left( {e(i)} \right)$ is a nonlinear function of $e(i) = d(i) - {W^T}(i)X(i)$ , which, for the MKRSL algorithm, is
\begin{equation}
f\left( {e(i)} \right) = \exp \left( {\lambda \left( {1 - {\kappa _\sigma }(e(i))} \right)} \right){\kappa _\sigma }(e(i))e(i)
\end{equation}
For the case $d(i) = {W_0}^TX(i) + v(i)$, the following relation holds {\cite{al2001adaptive}}:

\begin{equation}
\begin{aligned}
\textbf{E}\left[ {{{\left\| {\tilde W(i + 1)} \right\|}^2}} \right] =&  \textbf{E}\left[ {{{\left\| {\tilde W(i)} \right\|}^2}} \right] - 2\eta \textbf{E}\left[ {{e_a}(i)f(e(i))} \right] \\
&+ {\eta ^2}\textbf{E}\left[ {{{\left\| {X(i)} \right\|}^2}{f^2}(e(i))} \right]
\end{aligned}
\end{equation}
where $\tilde W(i) = {W_0} - W(i)$ is the \emph{weigh error vector} at iteration $i$, and ${e_a}(i) = {\tilde W^T}(i)X(i)$ is the a priori error. The relation (35) is a direct consequence of the \emph{energy conservation relation}{\cite{al2001adaptive,al2003transient}}.\\
1) \emph{Transient Behavior}
\par Based on (35) and under some assumptions, one can derive a dynamic equation to characterize the transient behavior of the \emph{weight error power} $\textbf{E}\left[ {{{\left\| {\tilde W(i)} \right\|}^2}} \right]$. Specifically, the following theorem holds {\cite{al2003transient}}.

\par \emph{Theorem 3}: Consider the adaptive filtering algorithm (33), where $e(i) = d(i) - {W^T}(i)X(i)$, and $d(i) = {W_0}^TX(i) + v(i)$. Assume that the noise process $\left\{ {v(i)} \right\}$ is i.i.d. and independent of the zero-mean input $X(i)$ and that the filter is long enough so that ${e_a}(i)$ is zero-mean Gaussian and that ${\left\| {X(i)} \right\|^2}$ and ${f^2}\left( {e(i)} \right)$ are uncorrelated. Then it holds that

\begin{equation}
\begin{aligned}
&\textbf{E}\left[ {\left\| {\tilde W(i + 1)} \right\|_\Sigma ^2} \right] = \textbf{E}\left[ {\left\| {\tilde W(i)} \right\|_\Sigma ^2} \right] \\
&- 2\eta {h_G}\left( {\textbf{E}\left[ {\left\| {\tilde W(i)} \right\|_{X(i){X^T}(i)}^2} \right]} \right) \times \textbf{E}\left[ {\left\| {\tilde W(i)} \right\|_{\Sigma X(i){X^T}(i)}^2} \right]\\
&{\rm{                                         }} + {\eta ^2}\textbf{E}\left[ {\left\| {X(i)} \right\|_\Sigma ^2} \right]{h_U}\left( {\textbf{E}\left[ {\left\| {\tilde W(i)} \right\|_{X(i){X^T}(i)}^2} \right]} \right)
\end{aligned}
\end{equation}
where $\left\| {\tilde W(i)} \right\|_\Sigma ^2 = {\tilde W^T}(i)\Sigma \tilde W(i)$, and the functions ${h_G}(.)$ and ${h_U}(.)$ are defined by
\begin{equation}
\begin{aligned}
&{h_G}\left( {\textbf{E}\left[ {e_a^2(i)} \right]} \right) = \frac{{\textbf{E}\left[ {{e_a}(i)f(e(i))} \right]}}{{\textbf{E}\left[ {e_a^2(i)} \right]}} \\
&{h_U}\left( {\textbf{E}\left[ {e_a^2(i)} \right]} \right) = \textbf{E}\left[ {{f^2}(e(i))} \right]
\end{aligned}
\end{equation}

\par \emph{Proof}: A detailed derivation can be found in {\cite{al2003transient}}.

\par For the MKRSL algorithm, $\forall x > 0$,the functions ${h_G}(x)$ and ${h_U}(x)$ can be expressed as
\begin{equation}
\begin{aligned}
{h_G}(x) =& \frac{1}{{\sqrt {2\pi {x^3}} }}\int_{ - \infty }^\infty  {\int_{ - \infty }^\infty  {y\exp \left( {\lambda \left( {1 - {\kappa _\sigma }(y + v)} \right)} \right)} } \times\\
&{\kappa _\sigma }(y + v)(y + v)\exp \left( { - \frac{{{y^2}}}{{2x}}} \right){p_v}(v)dydv\\
{h_U}(x) =& \frac{1}{{\sqrt {2\pi x} }}\int_{ - \infty }^\infty  {\int_{ - \infty }^\infty  {\exp \left( {2\lambda \left( {1 - {\kappa _\sigma }(y + v)} \right)} \right)} } \times\\
&{\kappa _{{\sigma  \mathord{\left/
 {\vphantom {\sigma  {\sqrt 2 }}} \right.
 \kern-\nulldelimiterspace} {\sqrt 2 }}}}(y + v){(y + v)^2}\exp \left( { - \frac{{{y^2}}}{{2x}}} \right){p_v}(v)dydv
\end{aligned}
\end{equation}
where ${p_v}(.)$ denotes the PDF of the noise $v(i)$. In general, there are no closed-form expressions for ${h_G}(.)$ and ${h_U}(.)$ . But the two functions can still be calculated by numerical integration.

\par \emph{Remark 6}: Using (36), one can construct the convergence curves of the weight error power. For example, if, in addition, the input sequence $\left\{ {X(i)} \right\}$ is i.i.d., with covariance matrix $\textbf{R} = \sigma _x^2\textbf{I}$, where $\textbf{I}$ denotes the identity matrix, we have

\begin{equation}
\begin{aligned}
&\textbf{E}\left[ {{{\left\| {\tilde W(i + 1)} \right\|}^2}} \right] = \textbf{E}\left[ {{{\left\| {\tilde W(i)} \right\|}^2}} \right] \\
&- 2\eta \sigma _x^2{h_G}\left( {\sigma _x^2\textbf{E}\left[ {{{\left\| {\tilde W(i)} \right\|}^2}} \right]} \right) \times \textbf{E}\left[ {{{\left\| {\tilde W(i)} \right\|}^2}} \right]\\
&{\rm{                                         }} + {\eta ^2}\sigma _x^2m{h_U}\left( {\sigma _x^2\textbf{E}\left[ {{{\left\| {\tilde W(i)} \right\|}^2}} \right]} \right)
\end{aligned}
\end{equation}
which is a recursion equation for generating the convergence curves of $\textbf{E}\left[ {{{\left\| {\tilde W(i)} \right\|}^2}} \right]$.\\
2) \emph{Steady-State Performance}
\par Let $S = \mathop {\lim }\limits_{i \to \infty } \textbf{E}\left[ {e_a^2(i)} \right]$ be the steady-state \emph{excess mean square error} (EMSE). According to {\cite{al2003transient}}, with the same setting of Theorem 3, the EMSE will be a positive solution of the following equation:

\begin{equation}
S = \frac{\eta }{2}Tr\left( \textbf{R} \right)\frac{{{h_U}(S)}}{{{h_G}(S)}}
\end{equation}
Since the functions ${h_G}(x)$ and ${h_U}(x)$ have no closed-form expressions in general, it is very difficult to solve the above equation. However, one can use a Taylor expansion method to obtain an approximate value of $S$. In this way, the following theorem holds.

\par \emph{Theorem  4}: Consider the adaptive filtering algorithm (33), where $e(i) = d(i) - {W^T}(i)X(i)$, and $d(i) = {W_0}^TX(i) + v(i)$. Assume that the noise process $\left\{ {v(i)} \right\}$ is zero-mean, i.i.d. and independent of the input $X(i)$ and that the \emph{a priori} error ${e_a}(i)$ is zero-mean and independent of the noise $v(i)$ and that ${e_a}(i)$ is relatively small at steady-state such that its third and higher-order terms are negligible. Then we have
\begin{equation}
S \approx \frac{{\eta Tr(\textbf{R})\textbf{E}\left[ {{f^2}(v)} \right]}}{{2\textbf{E}\left[ {f'(v)} \right] - \eta Tr(\textbf{R})\textbf{E}\left[ {f(v)f''(v) + {{\left| {f'(v)} \right|}^2}} \right]}}
\end{equation}
where $f'(v)$ and $f''(v)$ are the first and second derivatives of $f(v)$.

\par \emph{Proof}: see {\cite{chen2014steady}}.  for a detailed derivation.

\par For the MKRSL algorithm, the derivatives $f'(v)$ and $f''(v)$ are respectively

\begin{equation}
f'(v) = exp(\lambda (1 - {\kappa _\sigma }(v))){\kappa _\sigma }(v)\left( {1 + \lambda \frac{{{v^2}}}{{{\sigma ^2}}}{\kappa _\sigma }(v) - \frac{{{v^2}}}{{{\sigma ^2}}}} \right)
\end{equation}

\begin{equation}
\begin{aligned}
&f''(v) = exp(\lambda (1 - {\kappa _\sigma }(v))){\kappa _\sigma }(v)\times\\
&\!\left( {\frac{{{\lambda ^2}{v^3}}}{{{\sigma ^4}}}{\kappa _{{\sigma  \mathord{\left/
 {\vphantom {\sigma  2}} \right.
 \kern-\nulldelimiterspace} 2}}}(v) \!+\! \frac{{3\lambda {\sigma ^2}v \!-\! 3\lambda {v^3}}}{{{\sigma ^4}}}{\kappa _\sigma }(v) \!+\! \frac{{{v^3} \!-\! 3v{\sigma ^2}}}{{{\sigma ^4}}}} \right)
\end{aligned}
\end{equation}

\par \emph{Remark 7}: Given a noise PDF ${p_v}(.)$, one can calculate the expectations $\textbf{E}\left[ {{f^2}(v)} \right]$ , $\textbf{E}\left[ {f'(v)} \right]$ and $\textbf{E}\left[ {f(v)f''(v) + {{\left| {f'(v)} \right|}^2}} \right]$ , usually by numerical integration, and then obtain an approximate value of $S$ by using (41).

\section{SIMULATION RESULTS}
In this section, simulation results are presented to confirm the theoretical analysis and demonstrate the performance of the proposed MKRSL algorithm.
\subsection{Verification of Theoretical Results }
First, we demonstrate the theoretical and simulated convergence curves (in terms of the weight error power) of the MKRSL algorithm with different parameter settings. In the simulation, the filter length is set at $m=20$, and the initial weight vector is a null vector. The input and noise are both zero-mean white Gaussian processes with unit variance. The theoretical convergence curves and simulated ones averaged over 1000 Monte Carlo runs are shown in Fig. 3. As one can see, the theoretical curves match very well with those obtained by simulations.

\begin{figure}[h\t]
\setlength{\abovecaptionskip}{0pt}
\setlength{\belowcaptionskip}{0pt}
\centering
\includegraphics[width=3.0in,height=2.4in]{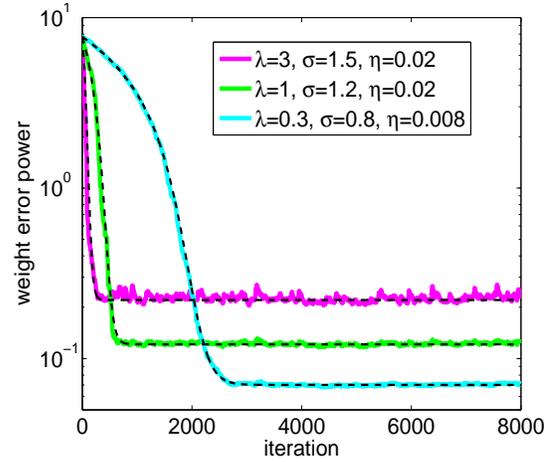}
\caption{Theoretical (black dashed) and simulated (solid) convergence curves with different parameter settings}
\label{fig3}
\end{figure}

\par We also illustrate the steady-state EMSEs. The unknown system and input are the same as in the previous simulation. The theoretical steady-state EMSEs calculated by (41) and simulated convergence curves (over 500 Monte Carlo runs) with different parameter settings are shown in Fig. 4, where the noise is assumed to be zero-mean Uniform distributed with unit variance. We can see that after transient stages the simulated curves will converge almost exactly toward the theoretical values. In addition, the theoretical and simulated steady-state EMSEs with different step-sizes and noise variances are shown in Fig. 5. To obtain the simulated steady-state EMSEs, we perform100 Monte Carlo simulations, and in each simulation, 200000 iterations are run to ensure the algorithm to achieve the steady-state, and the steady-state EMSEs are obtained as the averages over the last 10000 iterations. Again, simulation results agree very well with the theoretical predictions. To further confirm the theoretical results, we present in Table II the theoretical and simulated steady-state EMSEs for different noise distributions,  where Gaussian and Laplace noises are both zero-mean with unit variance, Binary noise is either -1.0 or 1.0 (each with probability 0.5), and Cauchy noise is distributed with PDF $p(v) = 1/[\pi (1 + {v^2})]$.
\begin{figure}[h\t]
\setlength{\abovecaptionskip}{0pt}
\setlength{\belowcaptionskip}{0pt}
\centering
\includegraphics[width=3.0in,height=2.4in]{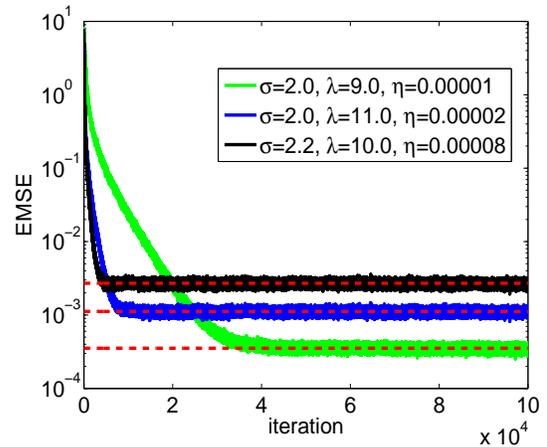}
\caption{Theoretical steady-state EMSEs (red dashed) and simulated convergence curves with different parameter settings.}
\label{fig4}
\end{figure}

\begin{figure*}[htbp]
\setlength{\abovecaptionskip}{0pt}
\setlength{\belowcaptionskip}{0pt}
\centering
\subfigure[]{
\includegraphics[width=3.0in,height=2.4in]{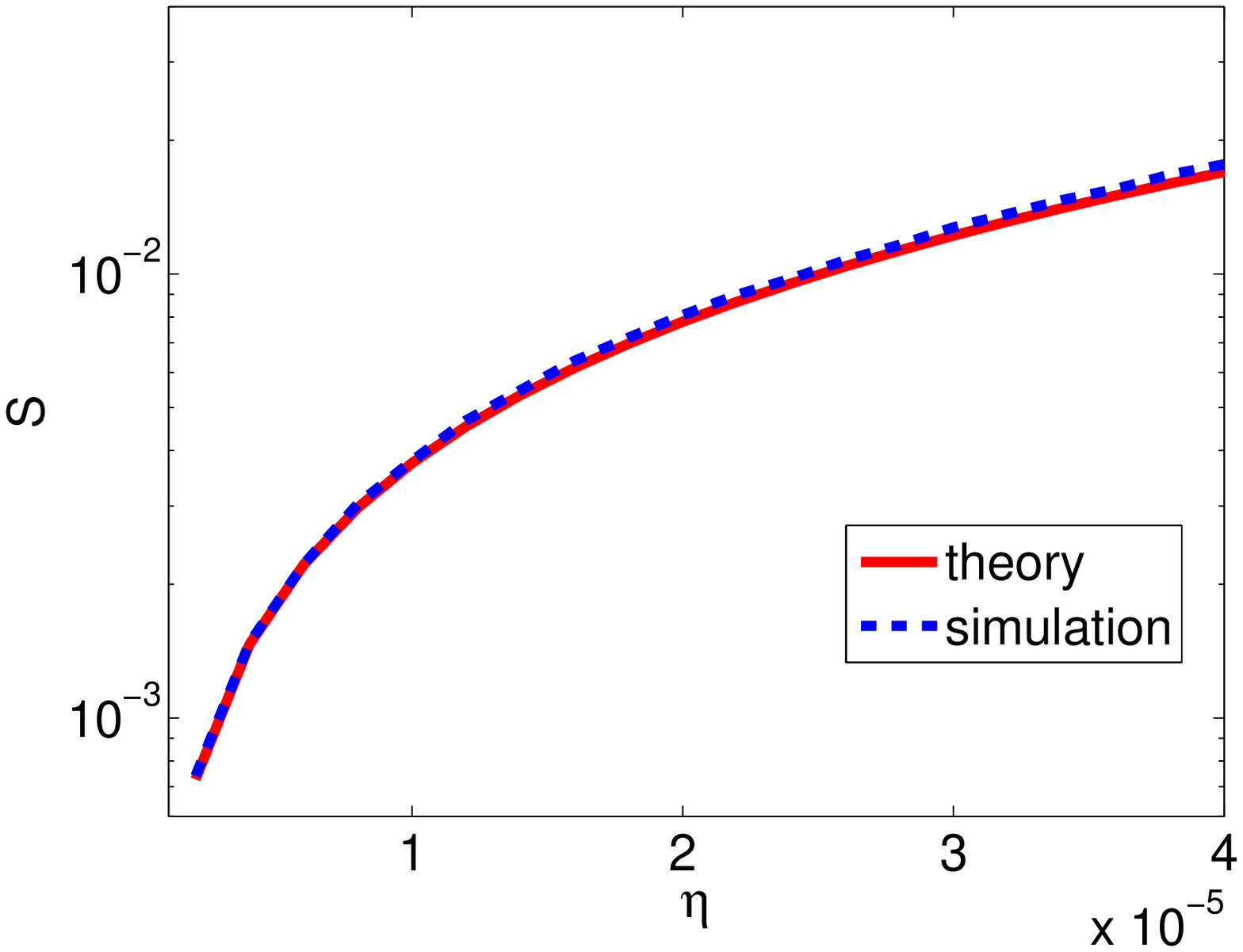}}
\subfigure[]{
\includegraphics[width=3.0in,height=2.4in]{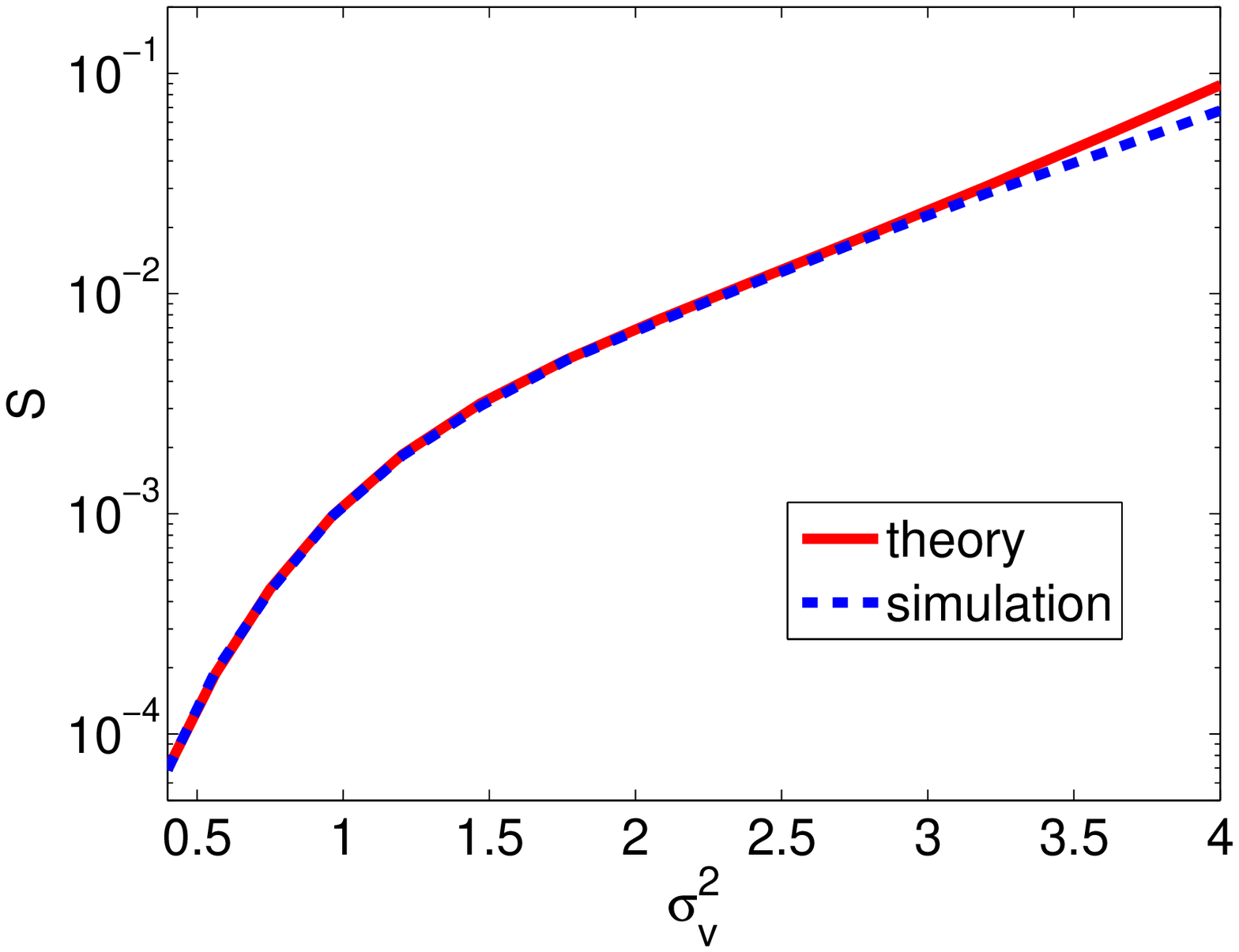}}
\caption{Theoretical and simulated steady-state EMSEs: (a) with different step-sizes ( $\lambda  = 8.0$, $\sigma=1.0$, $\sigma _v^2 = 1.0$); (b) with different noise variances ( $\lambda  = 8.0$, $\sigma=1.0$, $\eta  = {\rm{0}}{\rm{.000003}}$).}
\label{fig5}
\end{figure*}

\begin{table*}[htbp]
\setlength{\abovecaptionskip}{0pt}
\setlength{\belowcaptionskip}{0pt}
\centering
\caption{Theoretical and simulated steady-state EMSEs for different noise distributions}
 \begin{tabular}{cccc}
  \toprule
   Noise distribution&Parameter setting&Theory&Simulation\\
  \midrule
  Gaussian&$\lambda  = 8.0,\;\sigma  = 1.0,\;\eta  = {\rm{0}}{\rm{.000003}}$&0.0030&$0.0031 \pm 0.0005$\\
  Binary&$\lambda  = 9.0,\;\sigma  = 1.0,\;\eta  = {\rm{0}}{\rm{.000003}}$&0.000116&$0.000117 \pm 0.000018$\\
  Laplace&$\lambda  = 9.0,\;\sigma  = 1.0,\;\eta  = {\rm{0}}{\rm{.000002}}$&0.0065&$0.0064 \pm 0.0012$\\
  Cauuchy&$\lambda  = 8.0,\;\sigma  = 1.0,\;\eta  = {\rm{0}}{\rm{.000002}}$&0.0049&$0.0049 \pm 0.0013$\\

  \bottomrule
 \end{tabular}
\end{table*}

\subsection{Performance Comparison with Other Algorithms}
Next, we compare the performance of the MKRSL with that of the LMS, sign algorithm (SA) {\cite{mathews1987improved,shao1993signal}}, least mean mixed-norm (LMMN) algorithm {\cite{chambers1994least}}, least mean M-estimate (LMM) algorithm {\cite{zou2000least}} and GMCC ($\alpha=2, 4, 6$){\cite{chen2016generalized}}. The weight vector of the unknown system is assumed to be ${W_0} = {\left[ {0.1,0.2,0.3,0.4,0.5,0.4,0.3,0.2,0.1} \right]^T}$, and the initial weight vector of the adaptive filters is a null vector. The input signal is zero-mean Gaussian with variance 1.0 and the noise is assumed to be $v(i) = (1 - a(i))A(i) + a(i)B(i)$ , where $a(i)$ is a binary i.i.d. process with $\Pr \left\{ {a(i) = 1} \right\} = c$ and $\Pr \left\{ {a(i) = 0} \right\} = 1 - c$ ($0 \le c \le 1$ ), and $A(i)$ and$B(i)$ are two mutually independent noise processes (both independent of $a(i)$) with variances $\sigma {}_A^2$ and $\sigma _B^2$. Usually the variance $\sigma _B^2$ is much larger than the variance $\sigma {}_A^2$, thus $B(i)$ can represent large outliers. Without mentioned otherwise, $c$ is set to 0.06 and $B(i)$ is a white Gaussian process with $\sigma _B^2{\rm{ = }}15$. For the noise $A(i)$, we consider four cases: a) zero-mean Gaussian distribution with $\sigma {}_A^2{\rm{ = }}1.0$; b) Binary distribution with $Pr\{ A(i) =  - 1\}  = Pr\{ A(i) = 1\}  = 0.5$; c) Uniform distribution over $[ - \sqrt 5 ,\sqrt 5 ]$; and d) Sine wave noise $2\sin (\omega )$, with $\omega$ uniform distributed over $[ 0 ,2\pi ]$. Fig. 6 shows the convergence curves (in terms of the weight error power) averaged over 100 independent Monte Carlo runs. In the simulation, the step-sizes are chosen such that all the algorithms have almost the same initial convergence speed, and other parameters (if any) for each algorithm are experimentally selected to achieve desirable performance. The selected values of these parameters are included in the figures. It can be seen clearly that the MKRSL algorithm can significantly outperform other algorithms.

\begin{figure*}[htbp]
\setlength{\abovecaptionskip}{0pt}
\setlength{\belowcaptionskip}{0pt}
\centering
\subfigure[]{
\includegraphics[width=3.3in,height=2.4in]{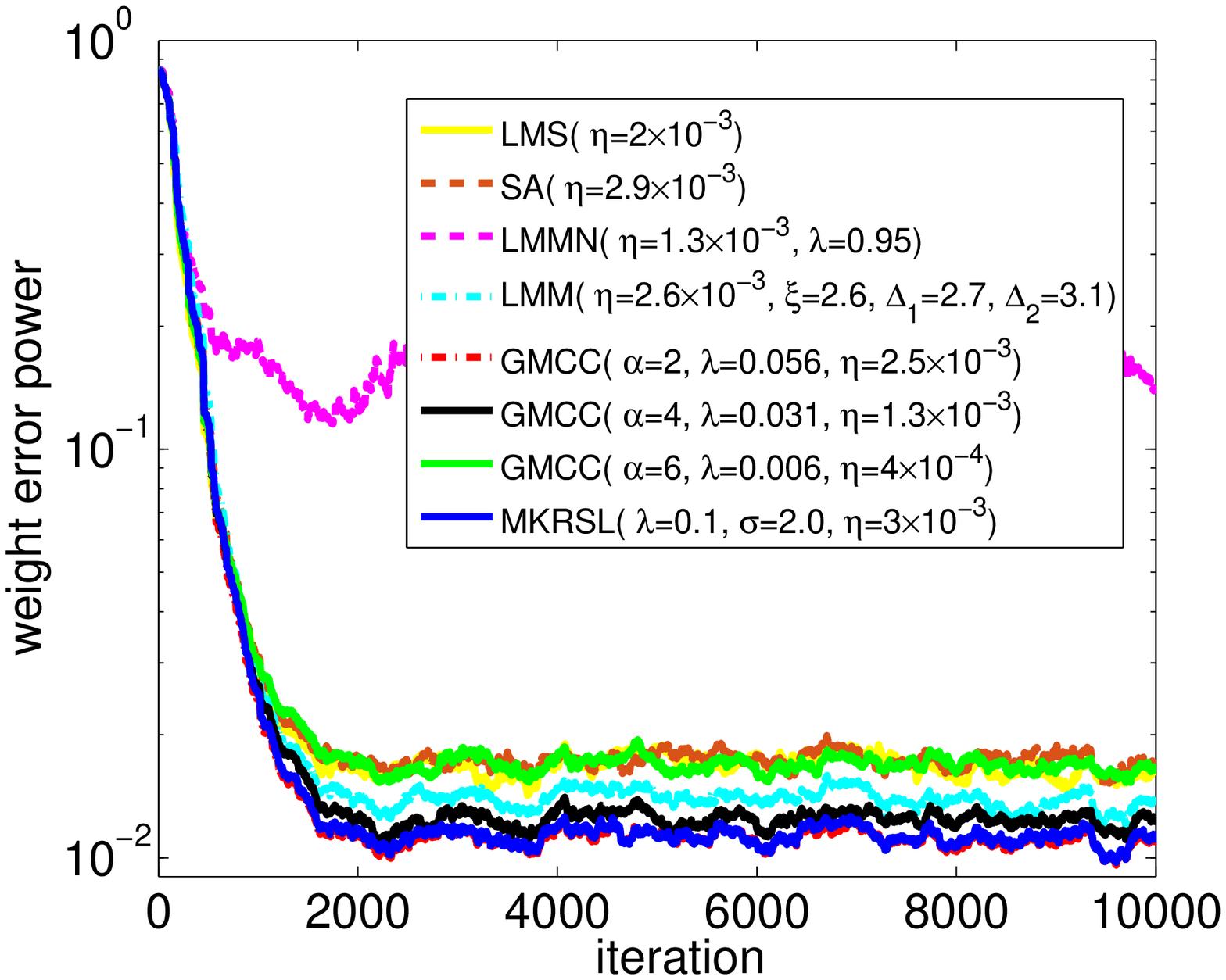}}
\subfigure[]{
\includegraphics[width=3.3in,height=2.4in]{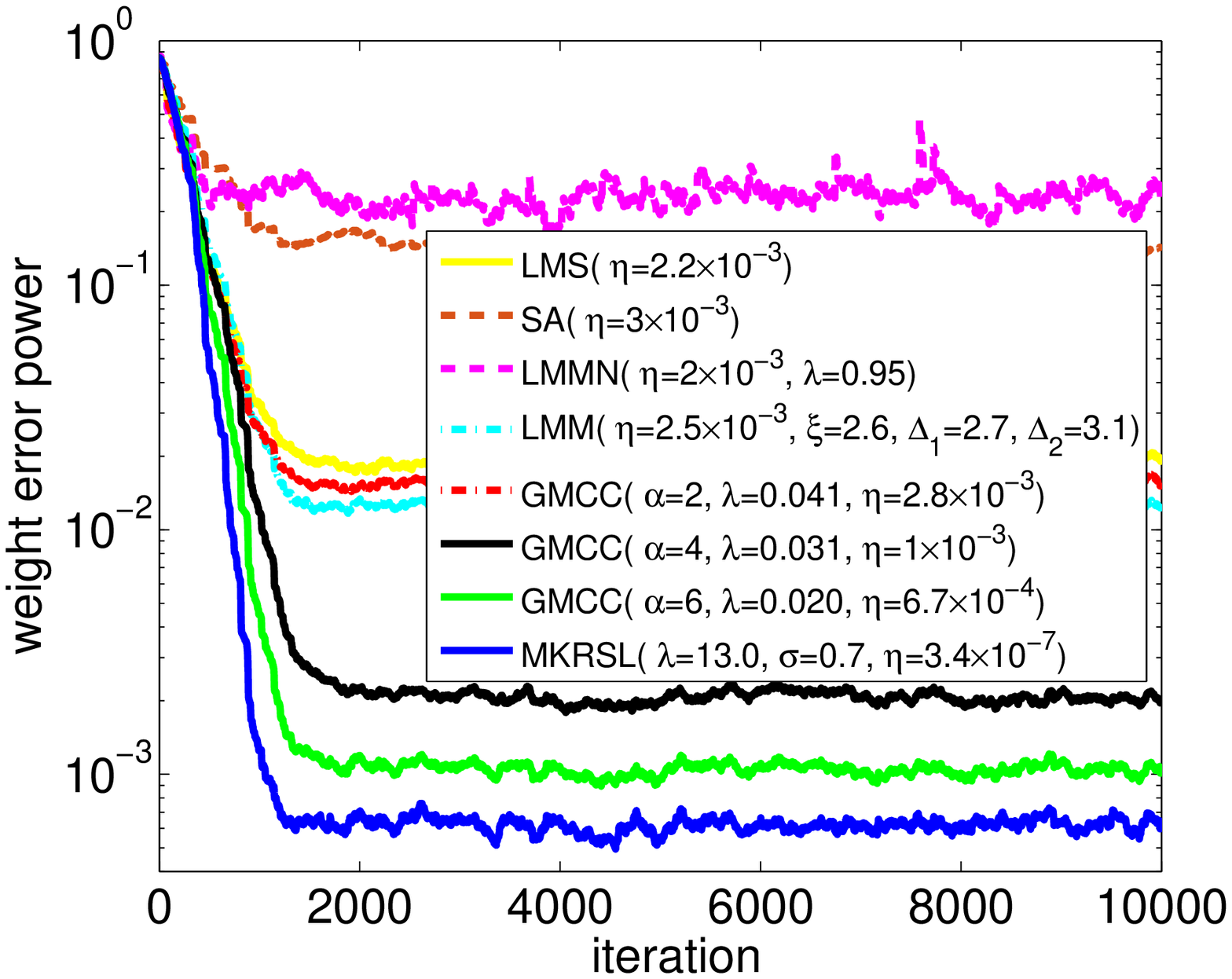}}
\subfigure[]{
\includegraphics[width=3.3in,height=2.4in]{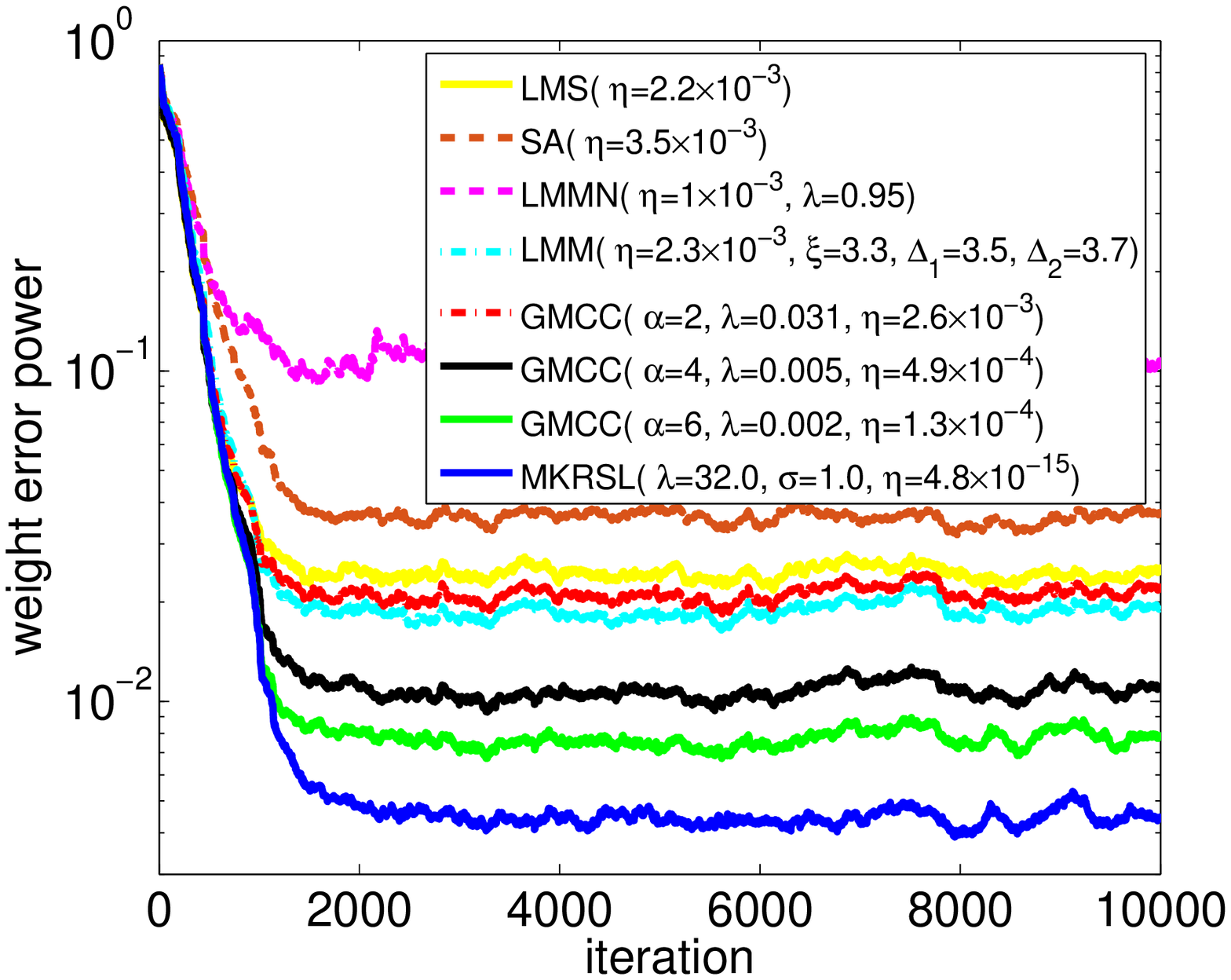}}
\subfigure[]{
\includegraphics[width=3.3in,height=2.4in]{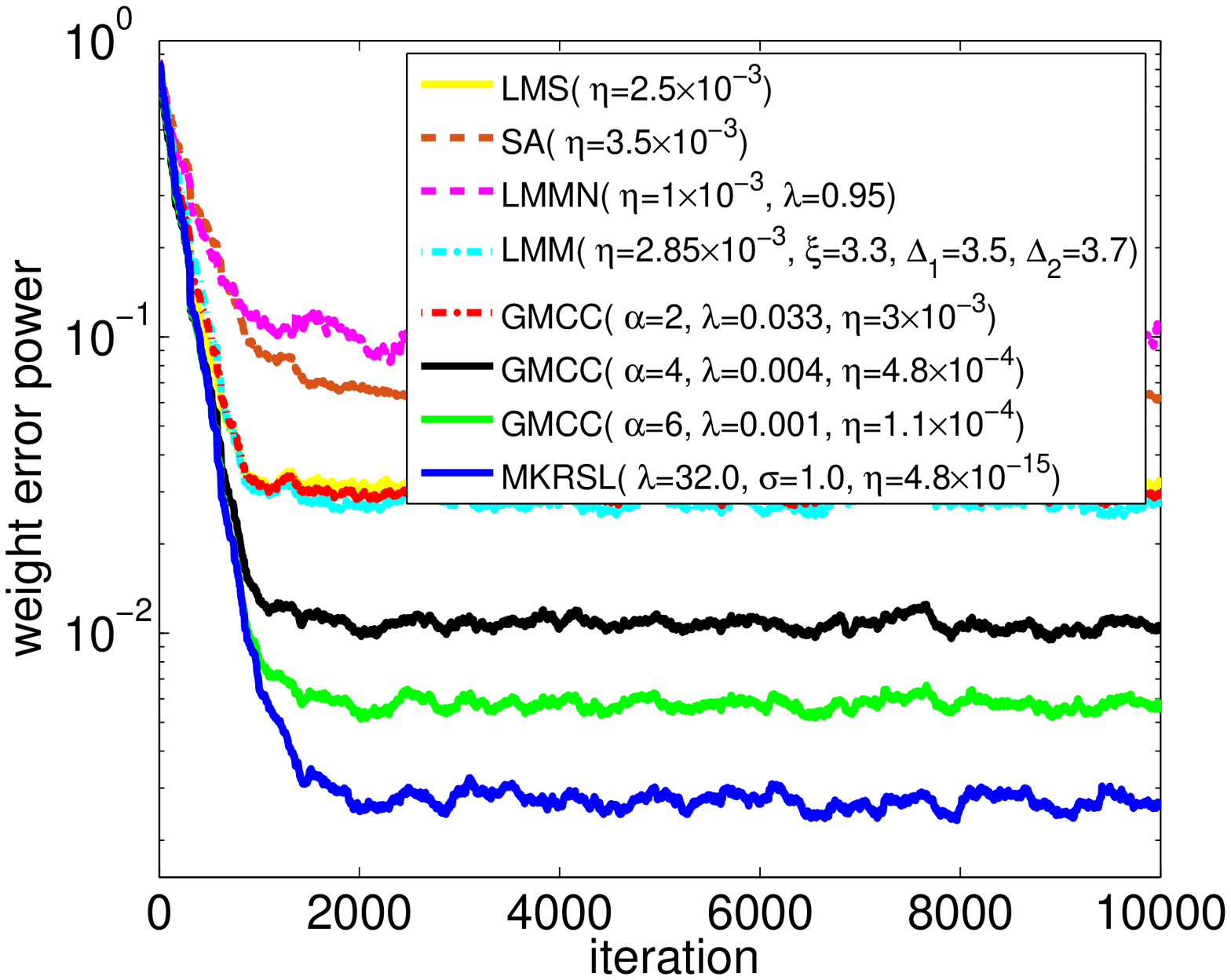}}
\caption{Convergence curves with different distributions of $A(i)$: (a) Gaussian; (b) Binary; (c) Uniform; (d) Sine wave}
\label{fig6}
\end{figure*}

\subsection{Effects of the Parameters $\lambda$ and $\sigma$}
Further, we show how the performance of the MKRSL will be influenced by the risk-sensitive parameter $\lambda$ and kernel bandwidth $\sigma$. With the same noise as in Fig.6 (c), the convergence curves of the MKRSL with different $\lambda$ and $\sigma$ are illustrated in Fig. 7. For each convergence curve, the step-size is chosen to achieve almost the same steady-state performance (Fig.7 (a)) or initial convergence speed (Fig.7 (b)). One can observe that both parameters have significant influence on the convergence behavior and desirable performance can be obtained only with appropriate parameter setting. How to determine an optimal value of $\lambda$ or $\sigma$ is however very involved and is left open in this work. In a practical application, the parameters $\lambda$ and $\sigma$ can be set manually or determined by \emph{trial and error} methods.


\subsection{Effects of the Outliers}
Finally, we demonstrate the robust performance of the MKRSL with different outlier variances ( $\sigma _B^2$) and occurrence frequencies ( $c$).With the same noise model as in Fig.6 (a), the steady-state weight error powers with different $\sigma _B^2$ and $c$ are plotted in Fig.8. From the simulation results we can observe: i) the MKRSL is very robust with respect to the amplitudes of outliers and its performance can even become better with the variance $\sigma _B^2$ increasing; ii) the algorithm is also robust with respect to the occurrence frequencies of outliers, and with $c$ increasing from 0\% to 30\%, the steady-state weight error power will increase very slightly (just from 0.0096 to 0.013).

\begin{figure*}[htbp]
\setlength{\abovecaptionskip}{0pt}
\setlength{\belowcaptionskip}{0pt}
\centering
\subfigure[]{
\includegraphics[width=3.0in,height=2.3in]{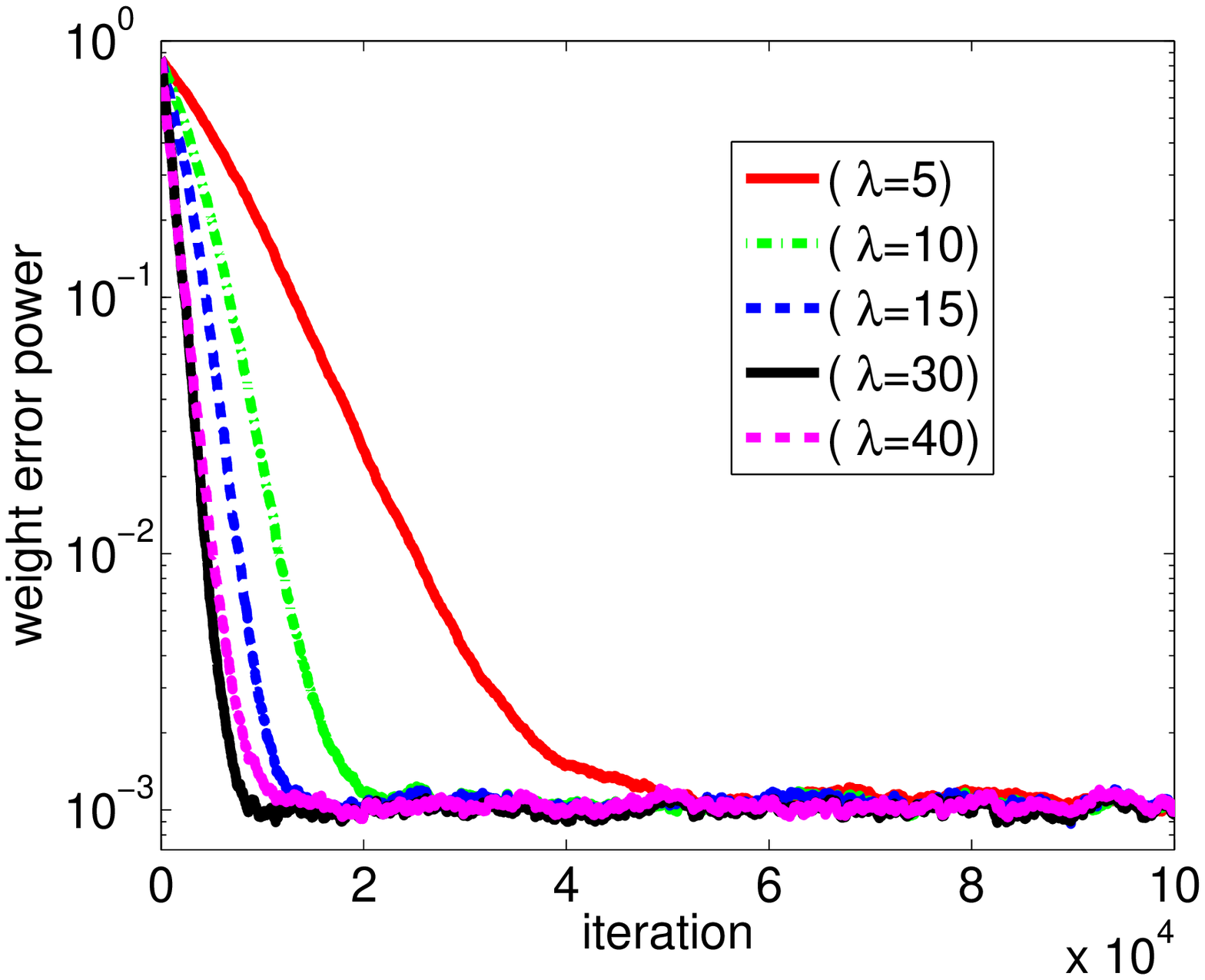}}
\subfigure[]{
\includegraphics[width=3.0in,height=2.3in]{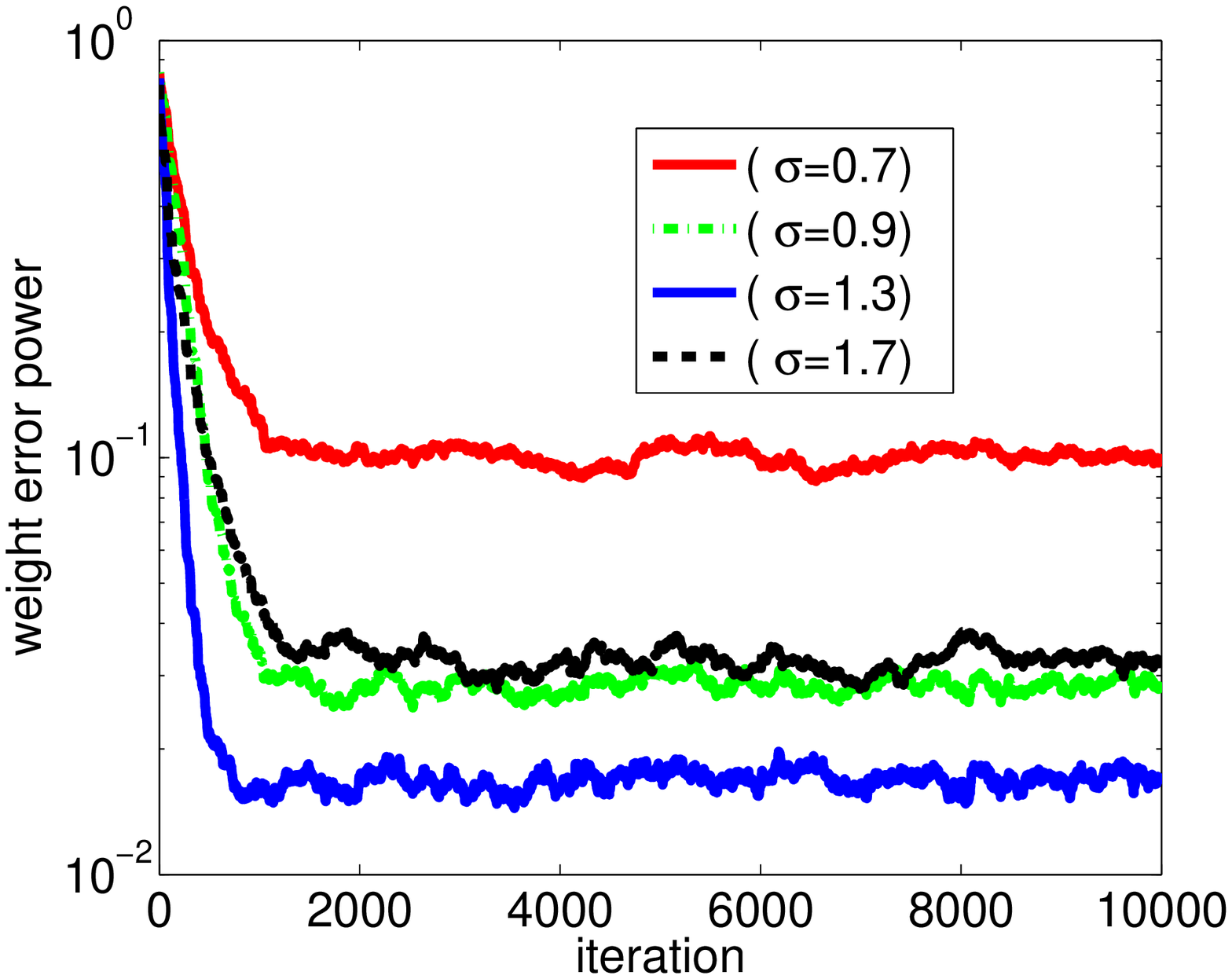}}
\caption{Convergence curves of MKRSL: (a) with different $\lambda$ ( $\sigma=1.0$); (b) with different $\sigma$ ( $\lambda=10$)}
\label{fig7}
\end{figure*}

\begin{figure}[h\t]
\setlength{\abovecaptionskip}{0pt}
\setlength{\belowcaptionskip}{0pt}
\centering
\includegraphics[width=3.0in,height=2.4in]{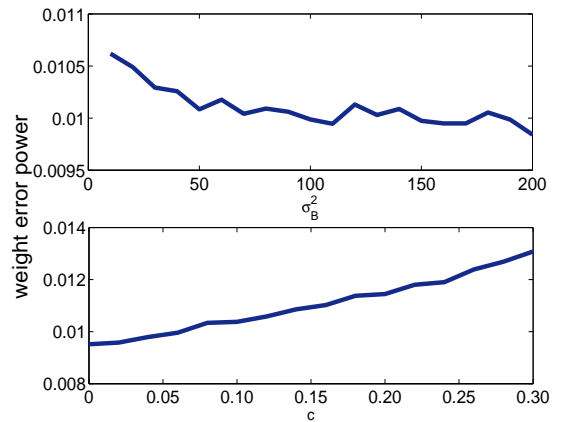}
\caption{Steady-state weight error powers with different outlier variances and occurrence frequencies}
\label{fig8}
\end{figure}

\section{CONCLUSION}
As a nonlinear similarity measure in kernel space, correntropy has been successfully applied in non-Gaussian signal processing and machine learning. To further improve the performance surface, we propose in this work a new similarity measure in kernel space, called the \emph{kernel risk-sensitive loss} (KRSL), which takes the same form as that of the traditional risk-sensitive loss, but defined in different spaces. Compared with correntropy, the KRSL can offer a more efficient performance surface that enables a gradient based method to achieve faster convergence speed and higher accuracy while still maintaining the robustness to outliers. Some important properties of the KRSL were presented. We applied the KRSL to adaptive filtering and investigated the robustness. Particularly, a robust adaptive filtering algorithm, namely the MKRSL algorithm, was derived and its mean square convergence performance was analyzed. Simulation results have confirmed the theoretical predictions and demonstrated that the new algorithm can achieve better convergence performance when compared with some existing algorithms.

\begin{appendices}
\section{PROOF OF THEOREM 2}
\emph{Proof}: Since ${W_{MKRSL}} = \arg \mathop {\min }\limits_W {J_{KRSL}}(W)$ , we have ${J_{KRSL}}({W_{MKRSL}}) \le {J_{KRSL}}({W_0})$ . To prove $\left| {{W_{MKRSL}} - {W_0}} \right| \le \xi $, it will suffice to prove ${J_{KRSL}}(W) > {J_{KRSL}}({W_o})$ for any $W$ satisfying $\left| {W - {W_0}} \right| > \xi $. Since $N > M > \frac{N}{2}$, we have $0 < \frac{{N - M}}{M} < 1$, and $0 < 1 - \frac{1}{\lambda }\log \left[ {\exp (\lambda ) - \frac{{N - M}}{M}\left( {\exp (\lambda ) - 1} \right)} \right] < 1$. As $\sigma  > \frac{{{\varepsilon _v}}}{{\sqrt { - 2\log \left\{ {1 - \frac{1}{\lambda }\log \left[ {\exp (\lambda ) - \frac{{N - M}}{M}\left( {\exp (\lambda ) - 1} \right)} \right]} \right\}} }}$, it follows easily that

\begin{equation}
0 \!<\! \frac{1}{\lambda }\!\log\!\! \left[ {\exp\! \left( \!{\lambda \!\left(\! {1 \!-\! \exp\! \left(\! { \!-\! \frac{{\varepsilon _v^2}}{{2{\sigma ^2}}}} \!\right)}\! \right)}\! \right) \!+\! \frac{{N \!-\!\! M}}{M}\!\left( {\exp \!(\lambda ) \!-\! 1} \!\right)} \!\right] \!\!<\! 1
\end{equation}
Further, if
\begin{equation}
\begin{aligned}
&\left| {W - {W_0}} \right| > \xi  = \frac{1}{c}\\
&\!\!\!\!\!\left(\!\!\! {\sqrt { \!\!-\! 2\!{\sigma ^2}\!\log\!\! \left(\!\! {1 \!\!\!-\!\! \frac{1}{\lambda }\!\log\!\! \left[\! {\exp\!\! \left( {\!\!\lambda\! \left(\!\! {1 \!\!\!-\!\! \exp\!\! \left( \!{ \!-\! \frac{{\varepsilon _v^2}}{{2{\sigma ^2}}}} \!\!\right)} \!\!\right)} \!\!\right) \!\!\!+\!\! \frac{{N \!\!\!-\!\!\! M}}{M}\!\!\left( {\!\exp \!(\!\lambda\! ) \!\!-\! 1} \!\right)} \!\right]} \!\right)}  \!\!\!+\!\!\! {\varepsilon _v}}\! \!\!\right)
\end{aligned}
\end{equation}
we have $\forall i \in I\left( {{\varepsilon _v}} \right)$,
\begin{equation}
\begin{aligned}
&\left| {e(i)} \right| = \left| {\left( {{W_0} - W} \right)X(i) + v(i)} \right|\\
&{\rm{        }} \ge \left| {{W_0} - W} \right| \times \left| {X(i)} \right| - \left| {v(i)} \right|\\
&{\rm{        }}\mathop  > \limits^{(a)} \xi c - {\varepsilon _v}=\\
&\!\!\!\!\!\! \sqrt { \!-\! 2{\sigma ^2}\!\log \!\!\left( \!\!{1 \!-\! \frac{1}{\lambda }\!\log\!\!\! \left[\! {\exp\!\! \left(\!\! {\lambda\!\! \left( \!\!{1 \!\!-\! \exp\!\! \left( { \!\!-\! \frac{{\varepsilon _v^2}}{{2{\sigma ^2}}}} \!\!\right)} \!\!\right)} \!\!\right) \!\!+\! \frac{{N \!\!-\! M}}{M}\!\left( {\exp\! (\lambda\! ) \!-\! 1} \!\right)}\! \right]} \!\right)}
\end{aligned}
\end{equation}
where (a) comes from the \emph{Assumption 2} and $\left| {W - {W_0}} \right| > \xi $ and $\left| {v(i)} \right| \le {\varepsilon _v}$. Thus $\forall i \in I\left( {{\varepsilon _v}} \right)$,
\begin{equation}
\begin{aligned}
&\exp \left( {\lambda \left( {1 - \exp \left( { - \frac{{{e^2}(i)}}{{2{\sigma ^2}}}} \right)} \right)} \right) - \exp (\lambda ) >\\
&\exp \left( {\lambda \left( {1 - \exp \left( { - \frac{{\varepsilon _v^2}}{{2{\sigma ^2}}}} \right)} \right)} \right) - \exp (\lambda ) - \frac{{N - M}}{M}\left( {1 - \exp (\lambda )} \right)
\end{aligned}
\end{equation}
Then we have ${J_{KRSL}}(W) > {J_{KRSL}}({W_o})$ for any $W$ satisfying $\left| {W - {W_0}} \right| > \xi $, because

\begin{equation}
\begin{aligned}
&{J_{KRSL}}(W) - \frac{1}{\lambda }\exp (\lambda )\\
&= \frac{1}{{N\lambda }}\sum\limits_{i = 1}^N {\left\{ {\exp \left( {\lambda \left( {1 - \exp \left( { - \frac{{{e^2}(i)}}{{2{\sigma ^2}}}} \right)} \right)} \right) - \exp (\lambda )} \right\}} \\
&> \!\!\frac{1}{{N\lambda }}\!\!\sum\limits_{i \in I\left( {{\varepsilon _v}} \right)} \!\!\!\!{\!\left\{ {\!\!\exp\!\! \left( {\!\!\lambda\!\! \left(\!\! {1 \!\!-\! \exp \!\!\left( { \!\!-\! \frac{{\varepsilon _v^2}}{{2{\sigma ^2}}}} \!\!\right)} \!\!\right)} \!\!\right) \!\!-\! \!\exp \!(\!\lambda ) \!\!-\! \frac{{N \!\!-\!\! M}}{M}\!\left( {1 \!\!-\! \exp \!(\lambda \!)} \!\right)} \!\!\right\}} \\
&+ \frac{1}{{N\lambda }}\sum\limits_{i \notin I\left( {{\varepsilon _v}} \right)} {\left\{ {\exp \left( {\lambda \left( {1 - \exp \left( { - \frac{{{e^2}(i)}}{{2{\sigma ^2}}}} \right)} \right)} \right) - \exp (\lambda )} \right\}} \\
&\mathop  > \limits^{(b)} \!\!\frac{1}{{N\lambda }}\!\!\!\!\sum\limits_{i \in I\left( {{\varepsilon _v}} \right)}\!\!\!\! {\left\{ {\!\exp\!\! \left( {\!\!\lambda\!\! \left(\! {1 \!\!-\! \exp\!\! \left(\! { \!-\! \frac{{\varepsilon _v^2}}{{2{\sigma ^2}}}} \!\!\right)} \!\!\right)} \!\!\right) \!\!-\! \exp \!(\!\lambda\! ) \!\!-\! \frac{{N \!\!-\!\! M}}{M}\!\!\left( {1 \!\!-\! \exp (\!\lambda\! )} \!\right)} \!\!\right\}} \\
&+ \frac{1}{{N\lambda }}\left( {N - M} \right)\left( {1 - \exp (\lambda )} \right)\\
&= \frac{1}{{N\lambda }}\sum\limits_{i \in I\left( {{\varepsilon _v}} \right)} {\left\{ {\exp \left( {\lambda \left( {1 - \exp \left( { - \frac{{\varepsilon _v^2}}{{2{\sigma ^2}}}} \right)} \right)} \right) - \exp (\lambda )} \right\}} \\
&\mathop  \ge \limits^{(c)} \frac{1}{{N\lambda }}\sum\limits_{i \in I\left( {{\varepsilon _v}} \right)} {\left\{ {\exp \left( {\lambda \left( {1 - \exp \left( { - \frac{{{v^2}(i)}}{{2{\sigma ^2}}}} \right)} \right)} \right) - \exp (\lambda )} \right\}} \\
&\mathop  > \limits^{(d)} \frac{1}{{N\lambda }}\sum\limits_{i = 1}^N {\left\{ {\exp \left( {\lambda \left( {1 - \exp \left( { - \frac{{{v^2}(i)}}{{2{\sigma ^2}}}} \right)} \right)} \right) - \exp (\lambda )} \right\}} \\
&= {J_{KRSL}}({W_0}) - \frac{1}{\lambda }\exp (\lambda )
\end{aligned}
\end{equation}
where (b) comes from $\exp \left( {\lambda \left( {1 - \exp \left( { - \frac{{{e^2}(i)}}{{2{\sigma ^2}}}} \right)} \right)} \right) \ge 1$, (c) follows from ${\varepsilon _v} \ge \left| {v(i)} \right|$, $\forall i \in I\left( {{\varepsilon _v}} \right)$, and (d) is due to $\exp \left( {\lambda \left( {1 - \exp \left( { - \frac{{{v^2}(i)}}{{2{\sigma ^2}}}} \right)} \right)} \right) - \exp (\lambda ) < 0$. This completes the proof.

\end{appendices}

\bibliographystyle{unsrt}
\bibliography{KRSL}
\end{document}